\documentclass[sn-mathphys,iicol]{sn-jnl}

\usepackage{booktabs}
\usepackage{multirow}%
\usepackage{amsmath,amssymb,amsfonts}%
\usepackage{amsthm}%
\usepackage{mathrsfs}%
\usepackage[title]{appendix}%
\usepackage{xcolor}%
\usepackage{textcomp}%
\usepackage{manyfoot}%
\usepackage{booktabs}%
\usepackage{graphicx}%
\usepackage{algorithm}%
\graphicspath{{figs/}}
\usepackage{algorithmicx}%
\usepackage{algpseudocode}%
\usepackage{listings}%

\theoremstyle{thmstyleone}%
\theoremstyle{thmstyletwo}%

\theoremstyle{thmstylethree}%

\begin{document}

\title[Article Title]{Enhanced Data-driven Topology Design Methodology with Multi-level Mesh and Correlation-based Mutation for Stress-related Multi-objective Optimization}


\author*[1]{\fnm{Jun} \sur{Yang}}\email{yang$\_$jun2023@fuji.waseda.jp}

\author[1]{\fnm{Shintaro} \sur{Yamasaki}}

\affil*[1]{\orgdiv{Graduate School of Information, Production and Systems}, \orgname{Waseda University}, \orgaddress{\street{2-7 Hibikino, Wakamatsu, Kitakyushu}, \city{Fukuoka}, \postcode{808-0135}, \country{Japan}}}


\abstract{

Topology optimization (TO) serves as a widely applied structural design approach to tackle various engineering problems.
Nevertheless, sensitivity-based TO methods usually struggle with solving strongly nonlinear optimization problems. 
By leveraging high capacity of deep generative model, which is an influential machine learning technique, the sensitivity-free data-driven topology design (DDTD) methodology is regarded as an effective means of overcoming these issues.
The DDTD methodology depends on initial dataset with a certain regularity, making its results highly sensitive to initial dataset quality. 
This limits its effectiveness and generalizability, especially for optimization problems without priori information.
In this research, we proposed a multi-level mesh DDTD-based method with correlation-based mutation module to escape from the limitation of the quality of the initial dataset on the results and enhance computational efficiency.
The core is to employ a correlation-based mutation module to assign new geometric features with physical meaning to the generated data, while utilizing a multi-level mesh strategy to progressively enhance the refinement of the structural representation, thus avoiding the maintenance of a high degree-of-freedom (DOF) representation throughout the iterative process.
The proposed multi-level mesh DDTD-based method can be driven by a low quality initial dataset without the need for time-consuming construction of a specific dataset, thus significantly increasing generality and reducing application difficulty, while further lowering computational cost of DDTD methodology.
Various comparison experiments with the traditional sensitivity-based TO methods on stress-related strongly nonlinear problems demonstrate the generality and effectiveness of the proposed method.

}



\noindent


\maketitle

\section{Introduction}\label{sec1}

Topology optimization, as a popular research topic, is a structural design methodology aimed at designing structures with optimal performance to address specific optimization problems while satisfying engineering requirements and functional demands.
Since its introduction by \cite{bendsoe1988generating}, TO has been successfully employed across a broad spectrum of engineering applications.
However, as TO methods continue to evolve and find broader applications, some researchers have increasingly recognized the limitations of mainstream sensitivity-based methods in handling
strongly nonlinear problems. 
Specifically, the existence of multiple local optima within the solution space of strongly nonlinear problems can cause sensitivity-based methods to become trapped in local optima that are low performance from an engineering perspective, e.g., turbulent flow channel design (\cite{dilgen2018topology}) and compliant mechanism design considering maximum stress (\cite{de2020stress}).
While some researchers have proposed several approaches to enhance sensitivity-based methods for addressing strongly nonlinear problems, these solutions are typically implemented indirectly and may lead to issues such as a loss of accuracy.

With the proposal of various sensitivity-free methods (\cite{chapman1994genetic,wang2006enhanced,tai2007target,wu2010topology}), some researchers have demonstrated their advantages in solving strongly nonlinear problems. 
However, these methods also reveal their difficulty in finding optimal or satisfactory solutions when solving optimization problem with high DOF.
Therefore, solving optimization problems with high DOF representations using a sensitivity-free methodology is a highly effective yet challenging solution for dealing with strongly nonlinear problems.
With the rapid development of artificial intelligence (AI), some researchers have recognized that incorporating deep generative modeling (a type of AI) into sensitivity-free methods brings novel inspiration to the aforementioned challenges (\cite{kingma2013auto}).
Due to the high fitting and carrying capability of deep neural networks, deep generative models can extract features from training data and generate diverse material distributions with high DOF by sampling in a small-scale latent space.

From this standpoint, \cite{yamasaki2021data} proposed a sensitivity-free DDTD methodology to efficiently tackle strongly nonlinear problems with a high DOF using deep generative model. 
In their research, part of material distributions are selected from a specific pre-constructed initial dataset containing multiple material distributions with a high DOF based on the set optimization objectives, that is elite data.
The selected elite data (or elite material distributions) are utilized to train the variational autoencoder (VAE), where their features are compressed into a small-sized latent space.
Latent variables are then sampled from this space and mapped back to the original DOF through the decoder.
Owing to the properties of deep generative models, the newly generated material distributions are diverse while inheriting features from the training data.
The newly generated material distributions are merged into the training data. 
Subsequently, new elite material distributions are selected from the merged data.
They are then used as the inputs for the next iteratation of VAE training. 
By iterating through this process, the performance of elite material distributions is enhanced while preserving a representation with a high DOF.
Leveraging the ability to handle optimization problems with high DOF and its sensitivity-free nature, DDTD enables to address strongly nonlinear optimization problems that are challenging or even infeasible to solve by mainstream sensitivity-based TO methods and has been applied in various research fields (\cite{yaji2022data,kato2023tackling,kii2023data,Kii2024latent}).
However, the current DDTD-based methods are reliant exclusively on deep generative models to bring diversity to the initial dataset.

This over-reliance on deep generative models leads to variations in the initial dataset significantly impacting the performance of the generated data, which is referred to as the limitation of the quality of the initial dataset on the results (QID-limitation) in this paper.
The nature of deep generative models is to generate diverse new data through unsupervised learning while inheriting original features from the input data. 
As a result, deep generative models struggle to learn meaningful features from an irregular initial dataset. 
This limitation is the primary reason why the current DDTD-based methods require the construction of a specific, regular, well-performance initial dataset.
However, constructing the initial dataset, for example, by solving a pseudo-problem that approximates the original optimization problem but is easier to solve (\cite{yaji2022data}), is effective, but may shift the focus of the DDTD methodology towards the construction of the pseudo-problem itself.
It is worth noting that, compared to a low quality initial dataset, constructing a specific initial dataset with a certain regularity typically results in higher computational costs and lower generalization.
Furthermore, due to the presence of QID-limitation, the current DDTD-based methods are not suitable for optimization problems that lack sufficient a priori information or face challenges in constructing simple yet analogous pseudo-problems to generate the required specific initial dataset.
Additionally, in the iterative process of original DDTD methodology, the representation of structures consistently maintains high DOF.
Although this facilitates describing complex structural shapes and variations, high DOF typically entail larger solution spaces and higher computational costs, which is not conducive to finding near-optimal solutions quickly.

Therefore, we propose a multi-level mesh DDTD-based method incorporating correlation-based mutation module to escape the impact of the quality of the initial dataset on the results while effectively improving the computational efficiency.
The correlation-based mutation module assigns new geometric features with physical meaning to the generated data, thereby avoiding over-reliance on deep generative models. 
This significantly enhances the ability of DDTD methodology to find optimal solutions from irregular, low quality datasets.
The multi-level mesh strategy aims to improve computational efficiency by representing the structure with multiple level of DOF, rather than maintaining the same level of mesh complexity throughout the iteration process.
The iterative process, following the multi-level mesh strategy, is roughly divided into two phases: low-mesh phase and enhanced-mesh phase.
In the low-mesh phase, near-optimal material distributions are rapidly generated from the initial layout under given load and boundary conditions, resulting in a fast decrease in the optimization objectives.
In the enhanced-mesh phase, based on the near-optimal material distributions obtained in the low-mesh phase, the details of the material distribution are gradually refined until the optimal material distributions are achieved, with the optimization objectives exhibiting fine-tuned changes.
Leveraging the multi-level mesh strategy and the correlation-based mutation module, the proposed methodology effectively tackles the QID-limitation and enhances computational efficiency. 
This leads to significantly faster convergence and better results compared to the original DDTD methodology without multi-level mesh strategy and the correlation-based mutation module.
Additionally, with its sensitivity-free nature and high generality, the proposed method can achieve superior results than the sensitivity-based TO methods on the strongly nonlinear problems e.g., maximum stress minimization problem.

In the following, the related work is introduced in section \ref{sec2} and the details of the proposed method are described in section \ref{sec3} and its effectiveness is confirmed using numerical examples in section \ref{sec4}. Finally, conclusions are provided in section \ref{sec5}.

\section{Related Work}\label{sec2}

\subsection{Topology Optimization}

TO is a powerful tool for generating high-performance structural designs within a defined domain and is widely applied across various fields. The homogenization method by \cite{bendsoe1988generating} initially transformed TO into a microstructural optimization problem, while the SIMP method by \cite{bendsoe2003topology} later introduced element density as a design variable to enhance computational efficiency.

Alternative approaches such as evolutionary structural optimization (ESO) by \cite{xie1996evolutionary} achieves structural topology optimization by removing material from low-stress regions. Boundary-based methods like level-set (\cite{allaire2002level}), MMC (\cite{guo2014doing}), and MMV (\cite{zhang2017explicit}) focus on boundary evolution for deformation control. Additionally, topological derivative (\cite{novotny2012topological}) and phase field (\cite{takezawa2010shape}) methods have contributed to expanding TO methodologies.

It is worth noting that these mainstream TO methods are usually based on sensitivity analysis to update the solution.
However, this dependence on gradient information can confine these methods to local optimal solutions, making it challenging to solve strongly nonlinear problems effectively.
In contrast, sensitivity-free TO methods can circumvent this issue and exhibit greater generalization. However, they encounter difficulties when applied to topology optimization problems with a high DOF.
Therefore, in recent years, some researchers have turned to incorporating machine learning (ML) techniques into the TO field to address these challenges.

\subsection{Machine Learning in Topology Optimization}

In recent years, there has been a surge of research combining ML with TO, aiming to enhance solution quality and lower computational expense.
\cite{banga20183d} introduced a deep learning-based methodology utilizing an encoder-decoder framework to expedite the TO process. 
Their approach leverages neural networks to efficiently explore design spaces.
\cite{zhang2019deep} proposed a deep convolutional neural network with robust generalization capabilities for structural TO.
Their method aims to enhance the robustness and efficiency of optimization outcomes.
\cite{chandrasekhar2021tounn} demonstrated a direct application of neural networks for topology optimization. Their work illustrates the utilization of neural network activation functions to represent material densities within a specified design domain, thereby integrating machine learning directly into the optimization process.
\cite{zhang2021tonr} conducted a comprehensive investigation into the direct application of neural networks (NN) for TO. 
Their method centers on reparameterization, transforming the conventional update of design variables in TO into the modification of the network's parameters, thereby enabling more efficient optimization.

Furthermore, \cite{jeong2023physics} proposed an innovative framework for TO called physics-informed neural network-based TO (PINN). 
This approach utilizes an energy-based PINN as a substitute for traditional finite element analysis in structural TO.
PINN is utilized to accurately compute the displacement field, thus enhancing the optimization process.
However, the optimization problems commonly addressed by ML-based TO methods, such as compliance minimization, can be solved with comparable effectiveness to those handled by traditional sensitivity-based approaches.
Despite the enhanced efficiency and effectiveness of ML-based approaches, they still encounter challenges in solving optimization problems that are inherently difficult for sensitivity-based methods, such as strongly nonlinear problems.

As mentioned earlier, sensitivity-free TO methods are effective for tackling strongly nonlinear problems. 
However, they face substantial challenges when dealing with large-scale optimization problems that involve high DOF. 
With the progress and application of ML in the TO field, deep generative models have been recognized by some researchers as a promising solution to tackle these challenges.
\cite{guo2018indirect} proposed a novel TO approach based on an indirect design representation.
This method leverages a VAE to encode material distributions and incorporates a style transfer technique to reduce noise, facilitating efficient exploration of the design space and the identification of optimized structures.
The research highlighted the capability of deep generative models to address large-scale optimization problems by compressing data into a latent space.
In contrast, \cite{oh2019deep} proposed a framework that iteratively combines TO with generative models to explore new design possibilities. 
This method produces a diverse range of designs from a small set of initial design data, enhancing the diversity of potential solutions. 
Similarly, \cite{zhang20193d} used a VAE to examine the 3D shape of a glider for conceptual design and optimization purposes, showcasing the applicability of deep generative models in complex optimization problem.

\begin{figure*}[t]
\begin {center}
\includegraphics[width=0.95 \textwidth]{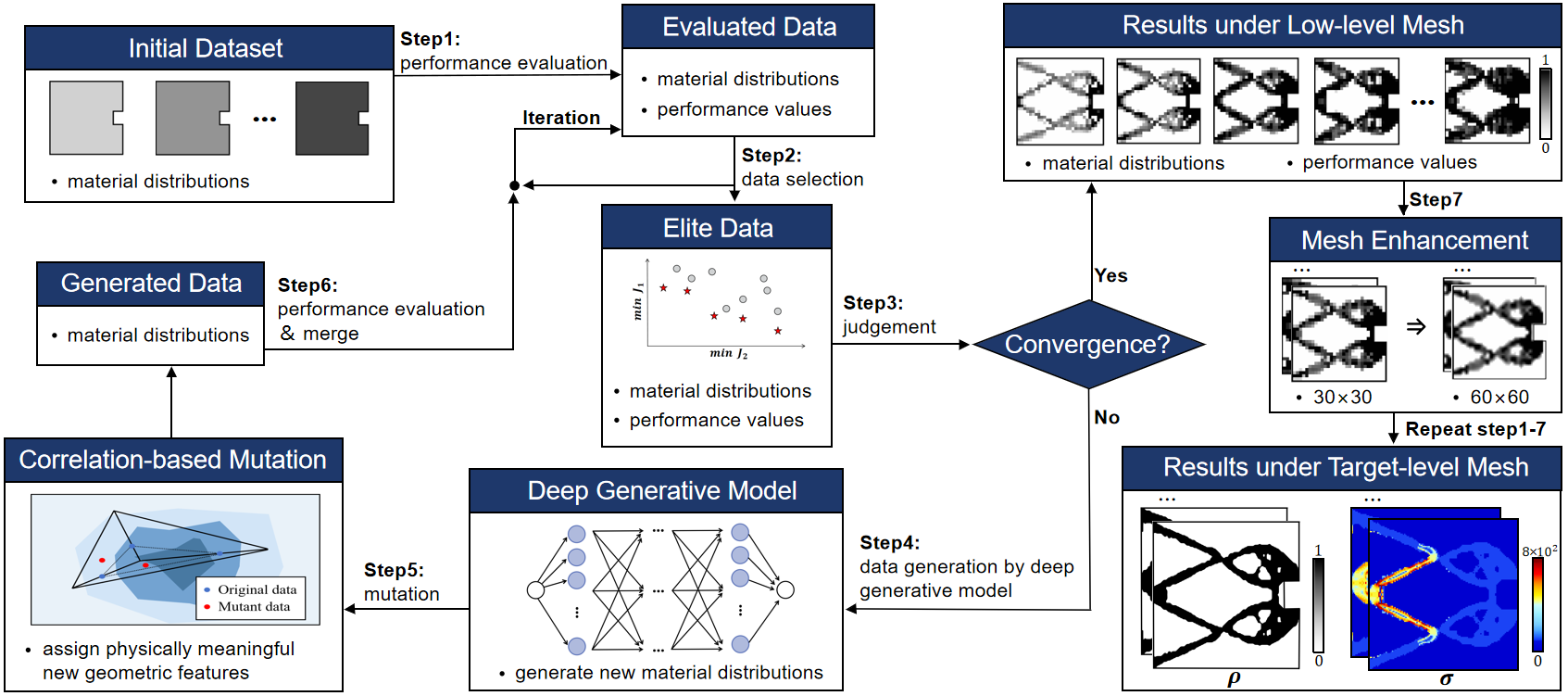}
\caption{Data process flow of the proposed multi-level mesh DDTD with correlation-based mutation module. }
\label{flowchart}
\end {center}
\end{figure*}

To tackle this challenge, \cite{yamasaki2021data} introduced a sensitivity-free DDTD methodology that integrates a policy to provide initial material distributions with a certain degree of regularity, ensuring the VAE can effectively capture meaningful features.
Furthermore, DDTD relies on high-performance material distributions for VAE training, distinguishing it from other approaches that use diverse material distributions as inputs.
With its sensitivity-free nature and capacity to solve large-scale problems, DDTD offers an effective solution to strongly nonlinear optimization.
Consequently, DDTD has found applications in various research fields, showcasing its versatility and efficacy.
\cite{yaji2022data} introduced a data-driven multifidelity topology design (MFTD) approach, which enables gradient-free optimization for complex problems with high DOF. They applied this method to forced convection heat transfer problems, demonstrating its effectiveness.
\cite{kato2023tackling} tackled a bi-objective problem focused on minimizing both the exact maximum stress and volume.
They employed a data-driven MFTD approach, integrating initial solutions derived from optimized designs obtained via gradient-based TO using the p-norm stress measure.

Furthermore, a series of other research (\cite{kii2023data,Kii2024latent}) have also focused on using data-driven DDTD to overcome current challenges in the field of TO. These studies highlight the potential and versatility of data-driven approaches in addressing complex optimization problems in various applications.

\section{Framework and implementation}\label{sec3}

\subsection{Fundamental formulation}

Let us consider a bounded and sufficiently regular domain $\Omega \subset \mathbb{R}^2$, which serves as the region for material optimization. 
The purpose of this research is to generate various material layouts within $\Omega$ that simultaneously satisfies multiple design objectives and constraints.
This problem can be formulated as:
\begin{equation}
\label{eq0}
\begin{array}{ll}
\underset{\rho}{\text{Minimize}} & \; \left[J_{1}(\rho), J_{2}(\rho), \cdots, J_{N_{\text{obj}}}(\rho) \right] \\ \\
\text{Subject to} & \; g_j(\rho) \leq 0, \;\; \text{for} \; j=1, 2, \dots, N_{\text{cns}}, \\ \\
& \;  \rho_{min} \leq \rho(\mathbf{x}) \leq 1 .
\end{array}
\end{equation}
where, $\rho(\mathbf{x})$ represents the design variable field, describing the material density at each point $\mathbf{x}$ in the domain. 
$J_{i}$ represents the $i$-th objective function, and $g_{j}(\rho)$ denotes the $j$-th constraint function.
$N_{\text{obj}}$, $N_{\text{cns}}$ are the numbers of $J$ and $g$, respectively.

To integrate the evaluations of the objective and constraint functions effectively, finite element analysis (FEA) is utilized. 
FEA simulates the physical behaviors within the design domain $\Omega$ by discretizing it into a mesh of finite elements. 
Each element's response to applied forces and deformations is computed based on its material properties, which vary according to the elemental density. 
This method allows us to evaluate how changes in material distribution influence the structural performance. Employing FEA facilitates a realistic representation of physical phenomena, ensuring that our optimization approach aligns with the practical constraints and behaviors of real-world materials.

\subsection{Data process flow }\label{sec32}

DDTD, being a sensitivity-free methodology, utilizes a deep generative model to generate diverse data that differ from the input data.
Nevertheless, DDTD's over-reliance on deep generative model leads to the appearance of QID-limitation, which can cause the results of current DDTD-based methods to be easily influenced by the quality of the initial dataset.
To address this limitation, we integrate the correlation-based mutation module into the DDTD methodology to assign geometric features with physical meaning on the original structures.
In this paper, the geometric feature with physical meaning is defined as a parameter-controlled component with certain randomness, its design being controlled by parameters such as position, shape and material density values.
It is intended to meet specific mechanical or functional requirements by changing the shape and topological relationships of the original structure. 
These geometric features bring diversity to a structure by selectively adjusting the densities of elements in specific areas.
In addition, while a high DOF representation is essential for accurately capturing the shape, topology, and variations of a structure, it is not advantageous for quickly obtaining a near-optimal structure during the early stages of iteration, and it demands significant computational cost.
Therefore, we propose a multi-level mesh strategy to represent the structure by setting multiple level DOF thereby reducing the computational cost instead of maintaining the same level of mesh complexity throughout the iteration process.

Figure~\ref{flowchart} shows the data process flow of the proposed multi-level mesh DDTD-based method and the details of each step are explained here.

\textbf{Initial data generation: } As the proposed methodology eliminates the need for constructing a specific initial dataset with a certain regularity, we employ straightforward ways to generate the initial dataset, aiming to minimize additional computational overhead.
This can be achieved, for example, by directly using the target volume fraction as the elemental density value.
It should be noted that the proposed method is not limited to low performance initial dataset; on the contrary, using high performance initial dataset can accelerate convergence.
However, preparing a high performance initial dataset typically results in a significant increase in computational cost and a decrease in generalizability. 
This is a key distinction that sets the proposed approach apart from the original DDTD methodology.

\begin{algorithm*}
\caption{Multi-level Mesh DDTD with Correlation-based Mutation Module}\label{algorithm}
\begin{algorithmic}[1]
\Require Set optimization problem.
\State Build VAE architecture and denoted as $V(*)$
\State Initialize the material distribution data: $\mathbf{X}_{\text{all}} $
\For{Low-level mesh to Target-level mesh}
\For{$i=0$ to $i_{\text{max}}$}
\State Evaluate performance of $\mathbf{X}_{\text{all}} $: $O_{\text{all}}$ \;
\State Select elite material distribution data: $\mathbf{X}$ $\Leftarrow$ $\mathbf{X}_{\text{all}}, O_{\text{all}}$\;
\If{the difference of outside area less than $1\times10^{-6}$ through 15 iterations }
\State break
\EndIf
\State Generate data via VAE: $\mathbf{X}_{\text{gen}} \Leftarrow V(\mathbf{X})$ \;
\State Assign geometric features via correlation-based mutation module: $\mathbf{X_{\text {mut }}} \Leftarrow \mathbf{X}_{\text{gen}}$\;
\If{the difference of outside area less than $1\times10^{-6}$ through 15 iterations }
\State Use heaviside function to update the mutated data: $\mathbf{X_{\text {nor }}} \Leftarrow \mathbf{X_{\text {mut }}} $ \;
\State Update material distribution data by merging: $\mathbf{X}_{\text{all}} \Leftarrow {\mathbf{X}}, \mathbf{X_{\text {norm }}}$\;
\Else
\State Update material distribution data by merging: $\mathbf{X}_{\text{all}} \Leftarrow {\mathbf{X}}, \mathbf{X_{\text {mut }}}$\;
\EndIf
\State $i \Leftarrow i+1$
\EndFor
\State Increase the DOFs of $\mathbf{X}_{\text{all}}$ \;
\EndFor
\State Obtain satisfactory elite material distribution data: $\mathbf{X}_{\text{all}}$\;
\end{algorithmic}
\end{algorithm*}

\textbf{Performance evaluation and data selection: } We evaluate the obtained material distribution data using the finite element method to determine the value of the optimization objectives within the set multi-objective optimization problem.
Subsequently, we perform the non-dominated sorting~\cite{deb2002fast} to obtain the rank-one material distributions as the elite data.
As previously mentioned, the process of proposed DDTD-based method is divided into two phases: the low-mesh phase and the enhanced-mesh phase. 
The low-mesh phase aims to obtain a near-optimal structure, utilizing a computationally efficient, low DOF structural representation to describe the material distribution.
Since both the generation and the evaluation process use the same number of DOF to describe the structures, they both have lower computational costs in the low-mesh phase, allowing the near-optimal solutions to be found more quickly due to the smaller solution space.
The enhanced-mesh phase aims to refine the structures with clear details, utilizing a computationally inefficient, high DOF structural representation to describe the material distributions.

\textbf{Data generation: } 
If the given convergence criterion is satisfied, we terminate the calculation and obtain the current elite data as the final results. 
Specifically, we define the convergence criterion as follows: if the change in the area outside of the elite solutions (an convergence indicator evaluating whole performance of the elite solution \citep{yamasaki2021data}, see Appendix) is below a predetermined threshold (usually $1\times10^{-6}$) for 15 consecutive iterations, or if the iteration count reaches the maximum iteration number, the optimization process is terminated.
The details of calculating the area outside of the elite solutions for each iteration are introduced in Appendix.
The generation process consists of two modules: the generation module based on deep generative model, and the correlation-based mutation module.
The former module trains a deep generative model using elite material distributions and produces generated data that inherits the original features of the input data while acquiring diverse.
The latter module mitigates the over-reliance of DDTD methodology on deep generative model, increases the sources of diverse features, and accelerates convergence by assigning new geometric features with physical meaning to the generated data.
After that, the generated data is updated using the heaviside function to reduce the number of greyscale elements in the material distributions.
It is important to note that this data update is typically carried out only within the enhanced-mesh phase, as the intermediate density plays a key role in the rapid generation of near-optimal structures within the low-mesh phase.
The architecture of the VAE and details are presented in Sec~\ref{sec34} and Sec~\ref{Generation and Correlation-based Mutation Module}, respectively.

\textbf{Mesh enhancement: } The number of DOF changes when transitioning from the low-mesh phase to the enhanced-mesh phase or continuing to enhance the mesh during the enhanced-mesh phase. 
In this paper, the low-mesh phase includes only a single mesh level, while the enhanced-mesh phase can contain multiple mesh levels, with the mesh having the highest DOFs referred to as target-level mesh.
Thus, the mesh enhancement algorithm is designed to generate an initial dataset for the enhanced-mesh phase at any DOF level from the results of the low-mesh phase or enhanced-mesh phase.
The details is presented in Sec~\ref{sec36}.

Through the above steps, we eliminate the QID-limitation existing in the current DDTD-based methods and reduce the computational cost by integrating the correlation-based mutation module and setting up a multi-level mesh structural representation.
To conclude the overview, we summarize the entire procedure in Algorithm~\ref{algorithm}.

\subsection{Variational Autoencoder}\label{sec34}

VAE is an unsupervised deep generative model capable of capturing meaningful features from input data while bringing diversity to the generated samples.
Unlike traditional deterministic autoencoders, VAEs introduce a latent variable model that regularizes the latent space, ensuring that generated samples follow a coherent and smooth distribution.

\begin{figure}[t]
\begin {center}
\includegraphics[width=0.45 \textwidth]{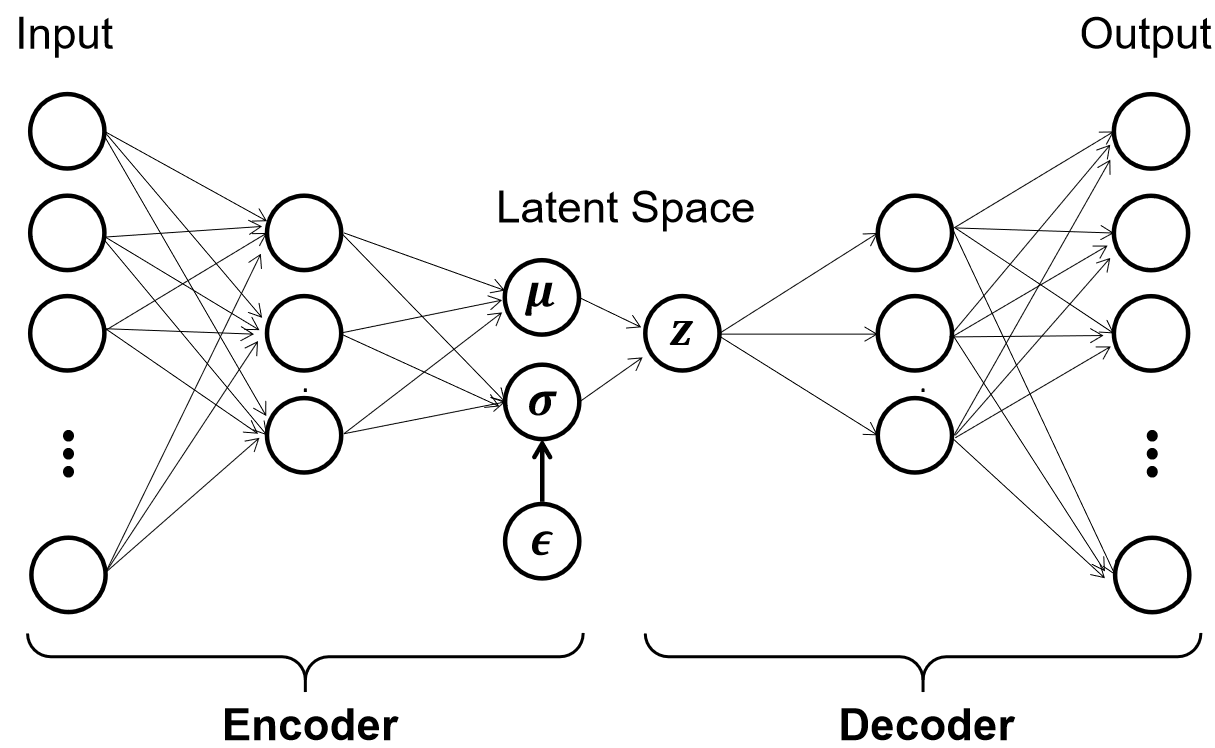}
\caption{Network architecture of the VAE}
\label{VAE}
\end {center}
\end{figure}

The VAE architecture consists of two fundamental components: an encoder and a decoder, as illustrated in Figure~\ref{VAE}. 
The encoder maps input data into a probabilistic latent space, parameterized by a mean vector $\boldsymbol{\mu} \in \mathbb{R}^{N_{\mathrm{ltn}}}$ ($N_{\mathrm{ltn}}$ represents the dimensionality of the latent space) and a variance vector $\boldsymbol{\sigma} \in \mathbb{R}^{N_{\mathrm{ltn}}}$ as following:
\begin{equation}
\label{eq7}
\mathbf{z}=\boldsymbol{\mu}+\boldsymbol{\sigma} \circ \epsilon
\end{equation}
where $\mathbf{z}$ are latent variables representing the latent space.
$\epsilon$ is a random vector sampled from a standard normal distribution, and $\circ$ denotes element-wise multiplication.
To ensure sufficient expressiveness while maintaining computational efficiency, the encoder consists of two fully connected hidden layers with 128 and 64 neurons, respectively. 
Subsequently, the decoder takes these latent variables and reconstructs them back into the dimension of input data, effectively generating new samples. 
The architecture of the encoder includes two fully connected hidden layers consisting of 64 and 128 neurons, respectively. 
By enforcing a probabilistic latent space, the VAE ensures that smooth interpolations between latent points produce meaningful variations of the data, rather than arbitrary noise.

The generative capability of VAE is driven by a composite loss function that considers both reconstruction accuracy and Kullback-Leibler (KL) divergence, as following:
\begin{equation}
L = L_{\mathrm{recon}} + \beta \cdot L_{\mathrm{KL}},
\label{eq8}
\end{equation}
where $L_{\mathrm{recon}}$ measures reconstruction accuracy, typically using the mean-squared error, while $L_{\mathrm{KL}}$ quantifies the KL divergence between the learned latent distribution and a prior Gaussian distribution. 
The hyperparameter $\beta$ regulates the trade-off between preserving fine details in generated samples and ensuring smooth organization of the latent space. The KL divergence is given by:
\begin{equation}
\label{eq9}
L_{\mathrm{KL}}=-\frac{1}{2} \sum_{i=1}^{N_{\mathrm{ltn}}}\left(1+\log \left(\sigma_{i}^{2}\right)-\mu_{i}^{2}-\sigma_{i}^{2}\right),
\end{equation}
where $\mu_{i}$ and $\sigma_{i}$ denote the $i$-th components of $\boldsymbol{\mu}$ and $\boldsymbol{\sigma}$, respectively.

By training the model to minimize this loss function, VAEs learn to generate new samples with diversity that adhere to the input data. 
The nature of the latent space ensures that even novel latent vectors sampled from the prior distribution correspond to realistic outputs, making VAE a powerful tool for controllable and interpretable generative modeling.

\subsection{Generation and Mutation Module}\label{Generation and Correlation-based Mutation Module}

\begin{figure*}[t]
\begin {center}
\includegraphics[width=1 \textwidth]{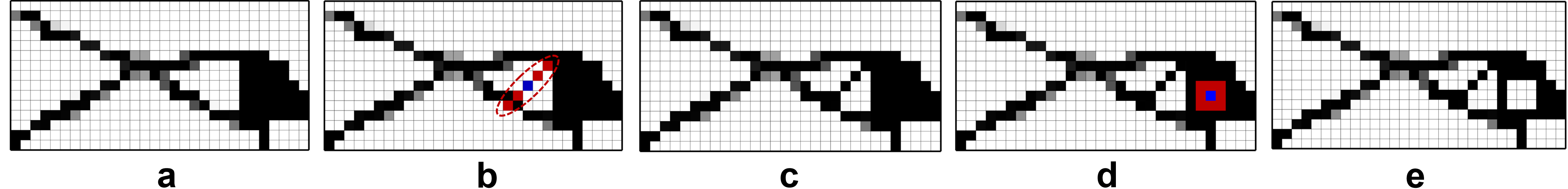}
\caption{The effect of $q$ value on the mutational geometric feature }
\label{mutation}
\end {center}
\end{figure*}

\begin{figure}[t]
\begin {center}
\includegraphics[width=0.45 \textwidth]{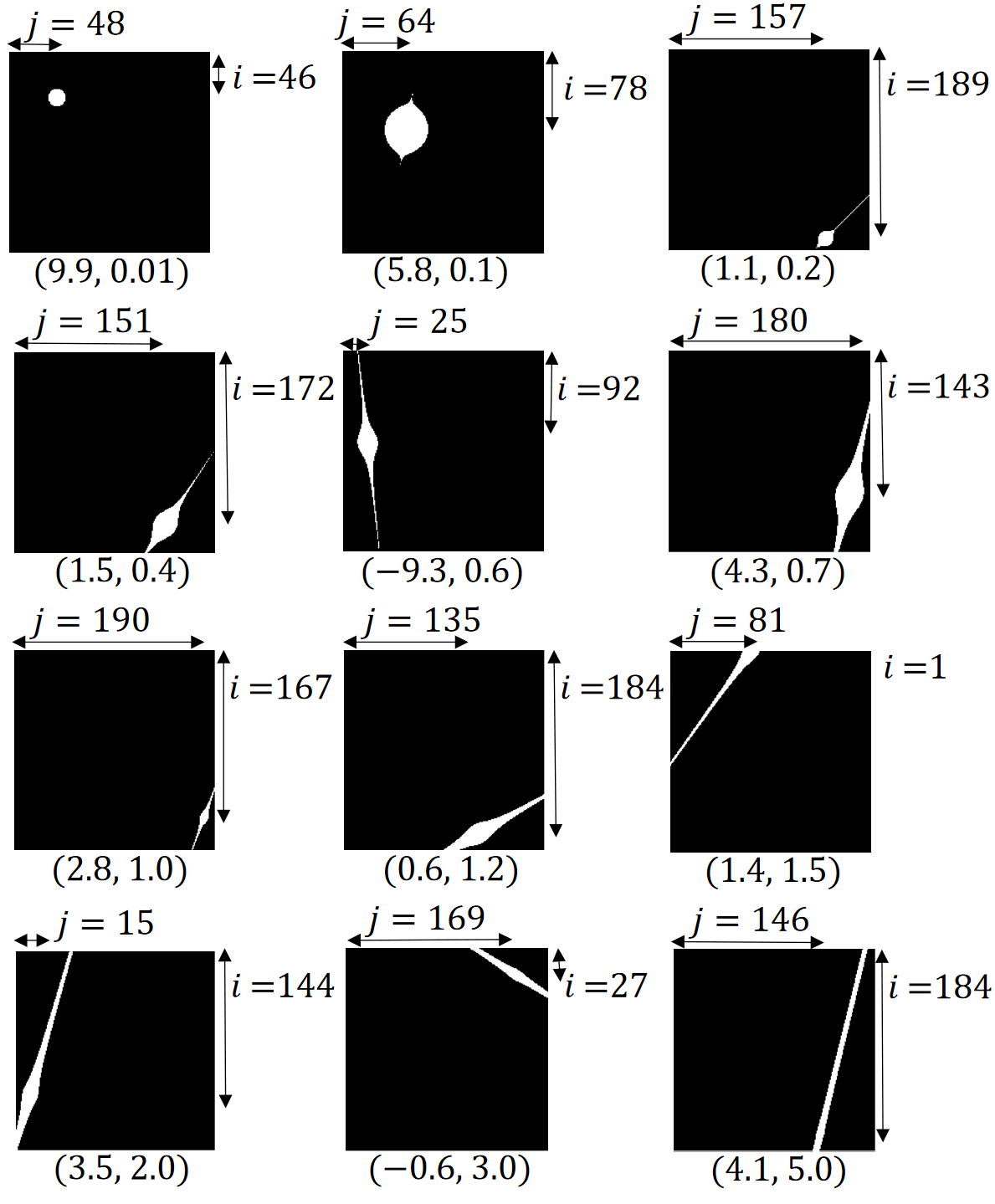}
\caption{Partial shape of the parameter-controlled components (white area) by random parameters. Here, $i, j$ denote the index of starting point in the $x, y$ direction ($n_x=n_y=200$), respectively. The left and right sides of the bracket indicate the $k_r$ and $q$ values, respectively}
\label{Fea}
\end {center}
\end{figure}

In this section, we introduce how to produce generated data with diversity based on the input data.
We note the material distributions of the elite data as $\mathbf{X} \in\mathbb{R}^{n \times m}$, as follows:
\begin{equation}
\mathbf{X} = \left[ \begin{array}{c}  \rho_1, ..., \rho_i, ..., \rho_m \end{array} \right],
\end{equation}
where $\boldsymbol{\rho}_{i} \in\mathbb{R}^{n \times 1}$ is the elemental density vector of the $i$-th material distribution, $m$ is the number of the material distributions, and $n$ is the number of DOF for each material distribution.

We denote the two modules of the generation process i.e. the generation module based on deep generative model, and the correlation-based mutation module as $G$, $M$ respectively.
In the former module, the input data $\mathbf{X}$ is fed into the VAE for training, resulting in generated data $G(\mathbf{X})$. 
This generated data inherits the original features of $\mathbf{X}$ while acquiring diversity by virtue of deep generative model's generative capacity.
The capabilities of deep generative models allow for modifications to the overall shape and topology of a structure by adjusting slight density variations in each element of the generated data.
However, this also poses a limitation as it may not effectively learn features from irregular data.
In original DDTD methodology, the quality of the initial dataset significantly impacts the final solution, i.e., the QID-limitation. 
This necessitates researchers to invest additional effort in constructing a specific, regular initial dataset for different optimization problems, e.g., constructing pseudo-optimization problems that are close to the original optimization problem but easier to solve.
Clearly, the DDTD-based methods face challenges in application when the optimization problem lacks sufficient a priori information or when constructing simple yet similar pseudo-problems to form the necessary initial dataset is not feasible.
The primary significance of the correlation-based mutation module lies in its ability to address the aforementioned challenges, i.e., to avoid over-reliance on deep generative model and add sources of diverse features.

In the latter module, new geometric features with physical meaning are assigned to the generated data by incorporating parameter-controlled components with certain randomness into the original structure.
The design of a parameter-controlled component consists with certain randomness mainly determined by two aspects: the starting point and the correlated area.
To illustrate how to derive a parameter-controlled component, let's consider a 2D example.
In this example, we use rectangular elements to discretize the design domain by the finite element method as shown in Figure~\ref{mutation}(a).
This design domain is divided into $n_{x}, n_{y}$ elements on the x, y-axis, respectively, and the number of elements is $n_{x}\times n_{y}$.
The starting point (blue element) is a randomly selected element among all the elements in the design domain, which serves as the first element contained in the parameter-controlled component as shown in Figure~\ref{mutation}(b).
The other elements contained in this component are determined by the correlation with the starting point.
The calculation of the correlation is related to the relative position between the start point and the other elements, and the position of the $d$-th element are defined as follows:
\begin{equation}
\mathbf{x_d} = (x_d, \, y_d),
\end{equation}
where $x_{d}, y_{d}$ are the coordinates of the x,y-axis of the center point of the $d$-th element (rectangular mesh), respectively.

After obtaining the position of the elements, the correlation between arbitrary element $\mathbf{x_d}$ of all the elements and the starting point $\mathbf{x_m}$ is calculated as follows:
\begin{equation}
\label{eq7}
A(\mathbf{x_d}, \, \mathbf{x_m}) = F^{q} \cdot\left\|\mathbf{x_m}-\mathbf{x_d}\right\|,
\end{equation}
where, $\mathbf{x_m}$ is the position of the starting point, $\left\| \cdot \right\| $ is the Euclidean distance between two points $\mathbf{x_m}$ and $\mathbf{x_d}$, $q$ is the weight parameter used as a scaling factor.
$F$ is utilized to control the degree of discretization between elements and is expressed as follows:
\begin{equation}
\label{eq8}
F=\frac{\left|k_{r} \cdot x_{m}-y_{m}+b_{0}\right|}{\sqrt{1+k_{r}^{2}}},
\end{equation}
where, $k_{r}$ is a random variable that ranges in $[-k_M, \, k_M]$, and $k_M$ is a constant value (usually $k_M=max(n_{x}/2, \, n_{y}/2)$). 
$x_{m}, y_{m}$ are the $x$ and $y$-axis coordinate values of $\mathbf{x_m}$, respectively. 
The constant term $b_{0}$ can be calculated by $b_{0}=y_m-k_{r}\cdot x_m$.
The random $n_m$ elements (usually $n_m<min(n_x, \, n_y)$) having the highest correlation with the starting point are selected as the correlated area.
By changing the value of scaling factor $q$ in Eq.\ref{eq7}, the shape of the parameter-controlled components (blue and red elements) can be altered to produce diverse geometric features as shown in  Figure~\ref{mutation}(b) and (d), in the settings of $q \rightarrow \infty$ and $q \rightarrow 0$, respectively.

The elemental densities $\rho_m$ of the parameter-controlled components are calculate as following:
\begin{equation}
\label{eq9}
\rho_m \sim U(\rho_{min},\, 1),
\end{equation}
where, $\rho_{min}$ is the minimum density. 
$U$ denotes a random value within the range.
Geometric features are assigned to the original structure to change its shape and topological relations by replacing the densities of the corresponding region with the parameter-controlled components as shown in Figure~\ref{mutation}(c) and (e) for $\rho_m=1,\, \rho_{min}$, respectively.
In contrast to random features without control on parameters such as position, shape and density, the design of geometric features are predictable within certain degrees as show in Figure~\ref{Fea}.
This can reduce the unpredictability due to random operation while may decreasing the ineffective material distribution. 
Compared to random features, geometric features with physical meaning usually possess properties such as structural continuity that satisfy various design objectives, and as a result, more substantial improvements can be introduced to the original structure to accelerate the convergence of the optimization process.

Through both modules, the generated data gains diversity to meet the requirements of various optimization problems.

The generated data is performed with density filter to eliminate the checkerboard problem.
\begin{equation}
\label{eq4}
\rho^{*}= \frac{W_{i}\left(\mathbf{x}_j\right) \rho_{j}} {\sum_{j=1}^{N} W_{i}\left(\mathbf{x}_j\right) \rho_{j}}
\end{equation}
where, $\rho^{*}$ is the filtered density, $\mathbf{x}_j,\rho_{j}$ are the elements surrounding the one element and the corresponding densities, respectively. 
$N$ is the number of surrounding the one element.
$W_{i} $ is the weight and $W_{i}\left(\mathbf{x}_j\right)=\max \left\{\mathrm{r}_{\min }-\mathrm{d}, 0\right\}$. 
$\mathrm{r}_{\min }$ is the minimum filter radius, which is a predefined value used to control the size of the filter range. 
It determines how the density values are smoothed or averaged over a range and is used to avoid numerical singularities or mesh dependency problems in the design.
$\mathrm{d}$ represents the Euclidean distance between any two points.

\begin{figure}[t]
\begin {center}
\includegraphics[width=0.5 \textwidth]{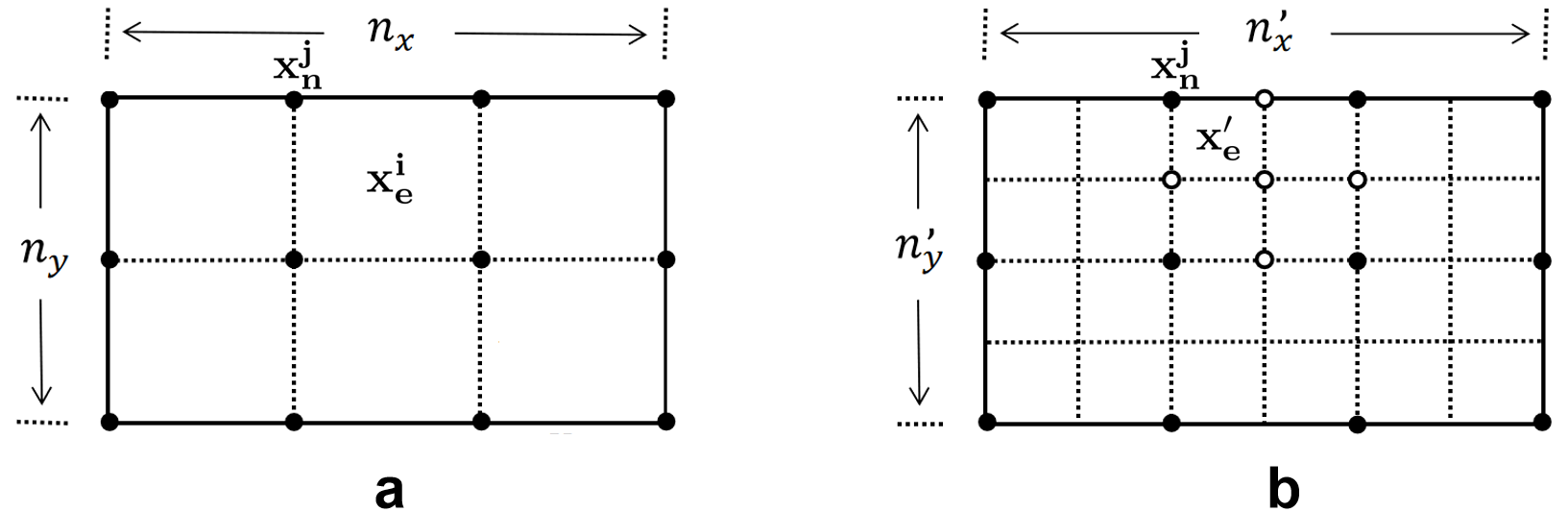}
\caption{Enhance mesh process: \textbf{a} low-level mesh \textbf{b} enhanced-level mesh}
\label{enhance mesh}
\end {center}
\end{figure}

In order to reduce the number of greyscale elements, we update the generated material distribution using the  heaviside function as following:
\begin{eqnarray}
& \hat{\rho}_{i} = \left\{\begin{array}{cl}
\rho_{min} & (\phi_i < -h) \\ \\
H(\phi_i) & (-h \leq \phi_i \leq h) \\ \\
1 & (h<\phi_i)
\end{array}\right. , \nonumber
\\
& \qquad\qquad\qquad\qquad \text{for} \; i=1, 2, \dots, n,
\end{eqnarray}
where $\hat{\rho}_{i}$ is the $i$-th components of the elemental density vector after the update, $\phi_i = 2 \rho_i - 1$ and $\rho_i$ is the $i$-th components of elemental density vector. $h$ is a constant parameter, and $H(\phi_i)$ is defined as follows:
\begin{equation}
H(\phi_i)=\frac{1}{2}+\frac{15}{16}\left(\frac{\phi_i}{h}\right)-\frac{5}{8}\left(\frac{\phi_i}{h}\right)^{3}+\frac{3}{16}\left(\frac{\phi_i}{h}\right)^{5}.
\end{equation}
In the numerical examples of this paper, we set $h$ as $0.01$, if we describe nothing.

\subsection{Mesh Enhancement}\label{sec36}

In this section, we describe how to enhance the material distributions from low-level mesh to high-level mesh.

As shown in Figure~\ref{enhance mesh}(a), the design domain is divided into $n_x,n_y$ via FEM and the coordinates of the nodes and elements (its centroids) are denoted as $\mathbf{x_n} (n=1,...,(n_x+1)\times(n_y+1))$ and $\mathbf{x_e} (e=1,...,n_x\times n_y)$, respectively.
The weight $w_j$ of $j$-th node at this level mesh are defined as follows:
\begin{equation}
\label{eq4}
w_j=\frac{\sum_{i=1}^{n_{w}} \rho_{i} \cdot\left\|\mathbf{x_n^j}-\mathbf{x_e^i}\right\|}{\sum_{i=1}^{n_{w}}\left\|\mathbf{x_n^j}-\mathbf{x_e^i}\right\|}
\end{equation}
where, $n_{w}$ is the number of elements surrounding $j$-th node (usually $n_{w}=4$), $\mathbf{x_n^j}$ is the coordinates of $j$-th node, $\mathbf{x_e^i}$ is the coordinates of $i$-th element, $\rho_{i}$ is corresponding density value of $i$-th element.

As shown in Figure~\ref{enhance mesh}(b), the design domain is divided into $n_x^{\prime},n_y^{\prime}$ after enhancing the mesh level.
The density $\rho^{\prime}$ of the newly divided element $\mathbf{x_e^{\prime}}$ can be calculated as follows:
\begin{equation}
\label{eq5}
\rho^{\prime}=\frac{\sum_{j=1}^{n_w} w_{j}\cdot\left\|\mathbf{x_n^j}-\mathbf{x_e^{\prime}}\right\|}{\sum_{i=1}^{n_{w}}\left\|\mathbf{x_n^j}-\mathbf{x_e^{\prime}}\right\|}
\end{equation}
where, $w_{j}$ is the weight of the surrounding nodes, $\mathbf{x_e^{\prime}}$ is the coordinates of newly divided element, $\mathbf{x_n^j}$ is the coordinates of $j$-th surrounding node.

With the above operations, low-level mesh data can be enhanced to high-level mesh as input data for generative model.
It should be noted that the enhancement of the mesh level only occurs at the end of the iteration at the current level mesh, and its aim is to provide input data for the iteration at the higher level mesh.

\section{Experiments}\label{sec4}

In this section, we conduct several numerical experiments to validate the effectiveness of the proposed method.
All experiments are implemented on a computer with Linux x86$\_$64 architecture and 128 cores.

\subsection{Problem setting}

\begin{figure*}[t]
\begin {center}
\includegraphics[width=1 \textwidth]{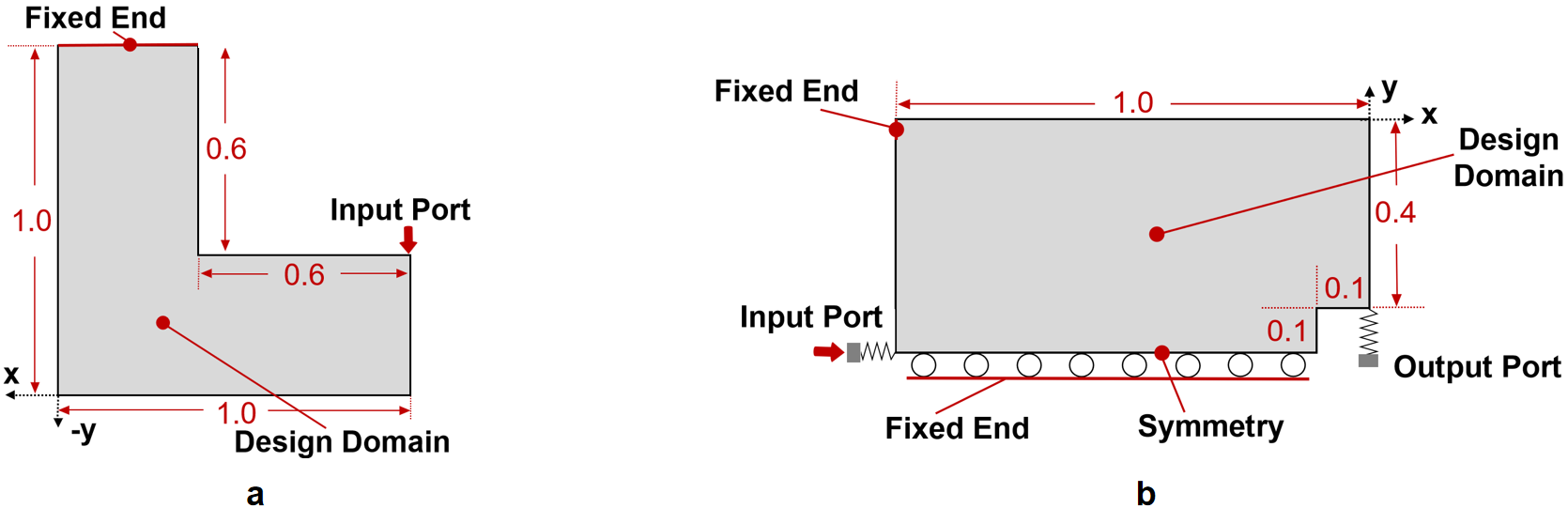}
\caption{Boundary conditions and design domains: \textbf{a} maximum von Mises stress (MVMS) minimization problem \textbf{b} compliant mechanism design problem considering stress}
\label{des}
\end {center}
\end{figure*}

In this paper, we focus on solving the following stress-related TO problems.
Sensitivity-based TO methods face significant challenges when addressing stress-related problems, primarily due to the strong nonlinearity of stress constraints and the localized nature of stress concentrations, which can lead to instability and difficulty in convergence. This issue is particularly pronounced when minimizing the maximum von Mises stress (MVMS), as directly computing its sensitivity is very challenging and prone to discontinuities and numerical instability, given that the maximum stress is a localized value. To overcome this, approximations such as the p-norm are often employed to smooth the local maximum stress into a global stress metric, facilitating smoother sensitivity calculations. While these approximations improve the problem’s solvability, they also introduce added computational complexity and potential error.

More specifically, in Sec~\ref{exam1}, we select the following objective functions in Eq~\ref{eqE1} to obtain L-shape structures whose MVMS is low:
\begin{equation}
\label{eqE1}
\text {Min: } \ \left\{\begin{array}{l}
J_1=\max \left(\sigma_{i}^{\text {mise }}\right) \qquad \quad  \ \qquad \qquad \\ \\ 
J_2 = \sum_{j=1}^{n} \rho_{ij} v_{j} \quad \quad \quad \quad 
\end{array}\right.
\end{equation}
\begin{equation}
\label{eq3}
\text { s.t.: }\ \left\{\begin{array}{l}
[K]\cdot \{u\}=\{f\} \ \\ \\
0 \textless \rho_{\min } \leqslant \rho_{ij} \leqslant 1 \quad\left(j=1,2, \cdots, n \right) \ \nonumber
\end{array}\right.
\end{equation}
where, $\sigma_{i}^{\text {mise }}$ is the stress distribution of the $i$-th structure in $\mathbf{X}$, $\rho_{ij}$ is the density of the $j$-th element in the $i$-th structure, $v_{j}$ is the volume fraction of the $j$-th element in the $i$-th structure, $n_e$ is the number of elemenets in one structure, $u$ is the displacement vector, $f$ is the loading. 
$K$ is the global stiffness matrix related to elemental stiffness $k_{i}$.
\begin{equation}
\label{eq3}
[K]=\sum_{i=1}^{n_{e}} k_{i}(\rho, \nu, e)
\end{equation}
where, $k_{i}$ is related to Young's modulus $e$, Poisson's ratio $\nu$ and density of element.
The von mises stress (VMS) of element can be calculated as following:
\begin{equation}
\sigma_{i}^{\text {mise }}=\sqrt{\sigma_{x}^{2}-\sigma_{x} \sigma_{y}+\sigma_{y}^{2}+3 \tau_{x y}^{2}}
\end{equation}
where, $\sigma_{x}, \sigma_{y}$ are the normal stress components in the $x$- and $y$-directions, respectively. 
$\tau_{x y}$ is the shear stress component in the $xy$-plane, and $\sigma_{j}=\left[\sigma_{x}, \sigma_{y},\tau_{x y} \right]$ can be calculated as following:
\begin{equation}
 \sigma_{j}=(\rho_{j})^{p} \cdot D_{0} \cdot B \cdot u_{j}
\end{equation}
where, $\sigma_{j}$ is the stress tensor for the $j$-th element, 
$(\rho_{j})^{p}$ is the density of $j$-th element, $p$ is a constant value (usually $p=0.5$). 
The introduction of the penalty term $(\rho_{j})^{p}$ enables that the contribution of stress from elements with lower density is reduced, effectively mitigating non-physical stress concentrations commonly observed at jagged boundaries \cite{deng2021efficient}.
According to the earlier definition, it consists of the stress components $\sigma_{x}, \sigma_{y},\tau_{x y}$.
The material's elastic matrix, which describes the material’s elastic properties. 
For isotropic materials, $B$ is the strain-displacement matrix.
$u_{j}$ is the nodal displacement vector for the $j$-th element, which contains the displacements of the element’s nodes under applied forces. 
$D_{0}\in\mathbb{R}^{3\times 3}$ is defined as follows:
\begin{equation}
D_{0}=\frac{E}{(1+\nu)(1-2 \nu)}\left[\begin{array}{ccc}
1-\nu & \nu & 0 \\
\nu & 1-\nu & 0 \\
0 & 0 & \frac{1-2 \nu}{2}
\end{array}\right]
\end{equation}
where, $E$ is the elastic modulus (usually $E=1$) and $\nu$ is the Poisson's ratio (usually $\nu=0.3$).

\begin{figure*}[t]
\begin {center}
\includegraphics[width=1 \textwidth]{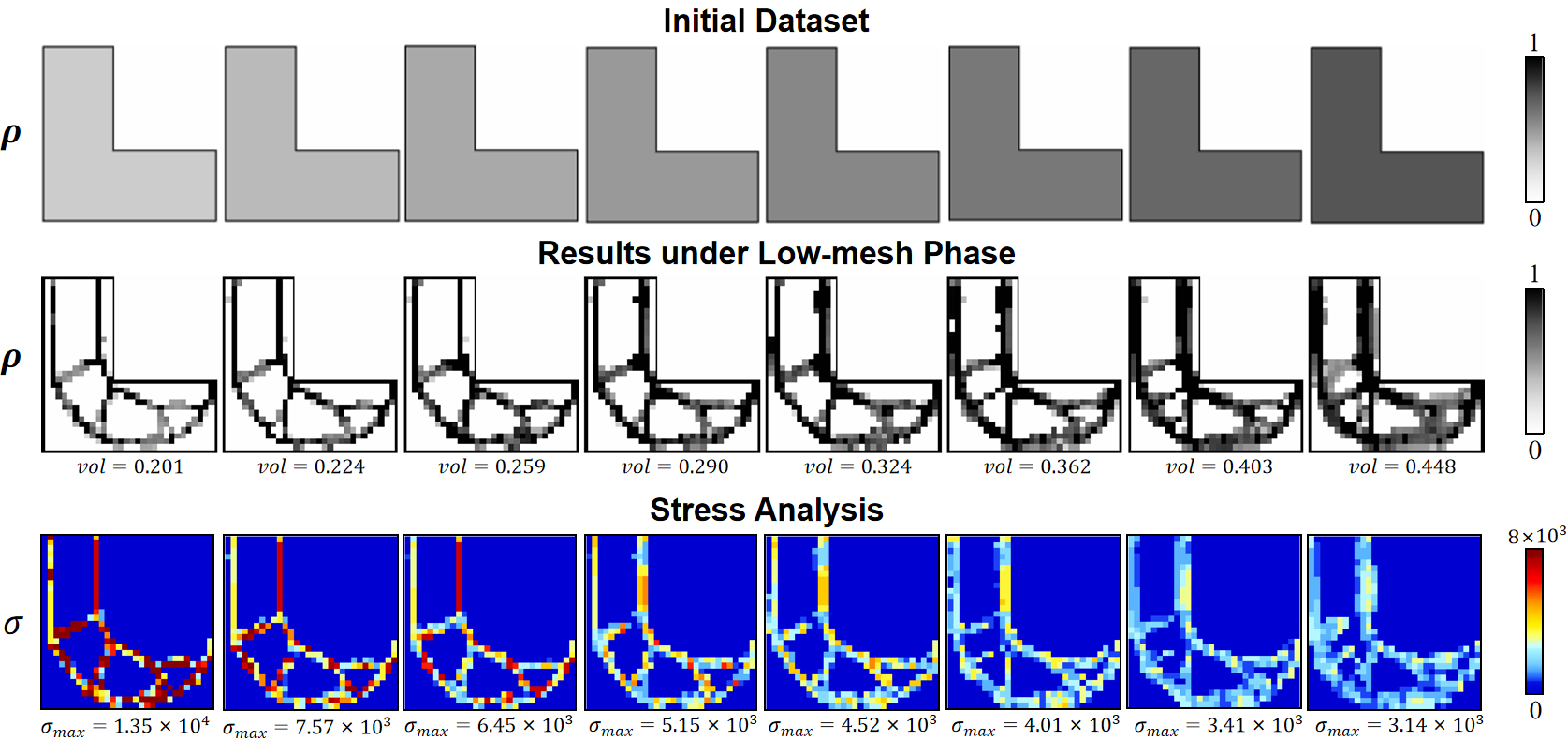}
\caption{Material distribution of elite data as well as stress analysis under low-level mesh: the first row is elite data at iteration~0, the second row is elite data at iteration~200, the third row is the corresponding stress distribution of elite data at iteration~200 }
\label{Experiment 1}
\end {center}
\end{figure*}

\begin{figure}[t]
\begin {center}
\includegraphics[width=0.4 \textwidth]{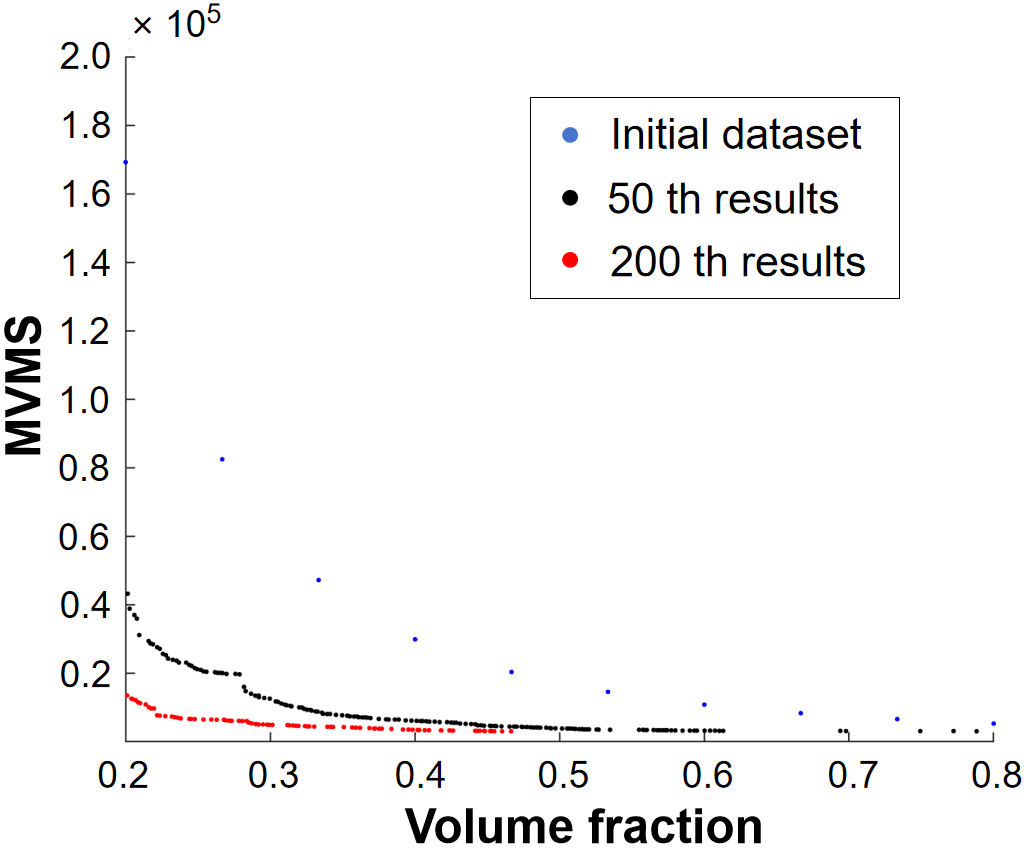}
\caption{Performances of elite material distributions: iteration~0 (blue), iteration~50 (black), and iteration~200 (red)}
\label{Improvement 1 Mesh 1}
\end {center}
\end{figure}

In Sec~\ref{exam2}, we select the followings as the objective functions in Eq~\ref{eq5} to obtain compliant mechanism structures:
\begin{equation}
\label{eq5}
\text {Min: } \ \left\{\begin{array}{l}
J_1=\max \left(\sigma_{i}^{\text {mise }}\right) \qquad \quad  \ \qquad \qquad \\ \\ 
J_2 = \sum_{j=1}^{n} \rho_{ij} v_{j} \quad \quad \quad \quad \\ \\
J_3 = -F_{\text{r}}
\end{array}\right.
\end{equation}
\begin{equation}
\text { s.t.: }\ \left\{\begin{array}{l}
[K]\cdot \{u\}=\{f\} \ \\ \\
0 \textless \rho_{\min } \leqslant \rho_{ij} \leqslant 1 \quad\left(j=1,2, \cdots, n\right) \ \nonumber
\end{array}\right.
\end{equation}
where, $F_{\text{r}}$ is the reaction force yielded by an artificial spring set on the output port. 
The negative sign of $F_{\text{r}}$ is needed to convert the maximization problem to a minimization problem.

As common settings, the following settings are adopted for the training of the VAE:
\begin{itemize}
    \item {mini-batch size}: $20$
    \item {number of epochs}: $400$
    \item {learning rate}: $1.0\times  10^{-4} $ 
    \item {number of latent variables, $N_{\text{ltn}}$}: $8$
    \item {weighting coefficient for KL divergence, $\beta$}: $1$
\end{itemize}
Furthermore, the latent crossover technique proposed by~\cite{Kii2024latent} is employed for sampling within the latent space.
The remaining parameters in the DDTD process are configured as follows:
\begin{itemize}
    \item {total number of iterations}: $50$
    \item {maximum number of elite data}: $400$
\end{itemize}

\subsection{Numerical example 1} \label{exam1}

First, L-shaped beam is selected as a test case to evaluate validity under the MVMS minimization optimization problem as shown in Figure~\ref{des}(a).
We establish two levels of mesh, meaning that the mesh only needs to be enhanced once.
The number of design variables of the low-mesh phase and the enhanced mesh phase are $576$ and $2304$, respectively.
The maximum number of iterations for the low-mesh phase and the enhanced mesh phase are $200$, $125$, respectively.
Furthermore, we imposed a volume fraction constraint ($J_2\in [0.2, 0.8]$) to ensure that the results fall within the target range.
For additional problem settings, a vertical load of $1500$ is applied at the tip of the L-shaped design domain, while the displacement is constrained at the fixed end. 
It should be noted that the loading is uniformly distributed on two nodes in the neighborhood at the right corner, which helps reduce stress concentrations typically caused by concentrated load. 
Young's modulus and Poisson's ratio are set to $1$ and $0.3$, respectively.

\begin{figure*}[t]
\begin {center}
\includegraphics[width=1 \textwidth]{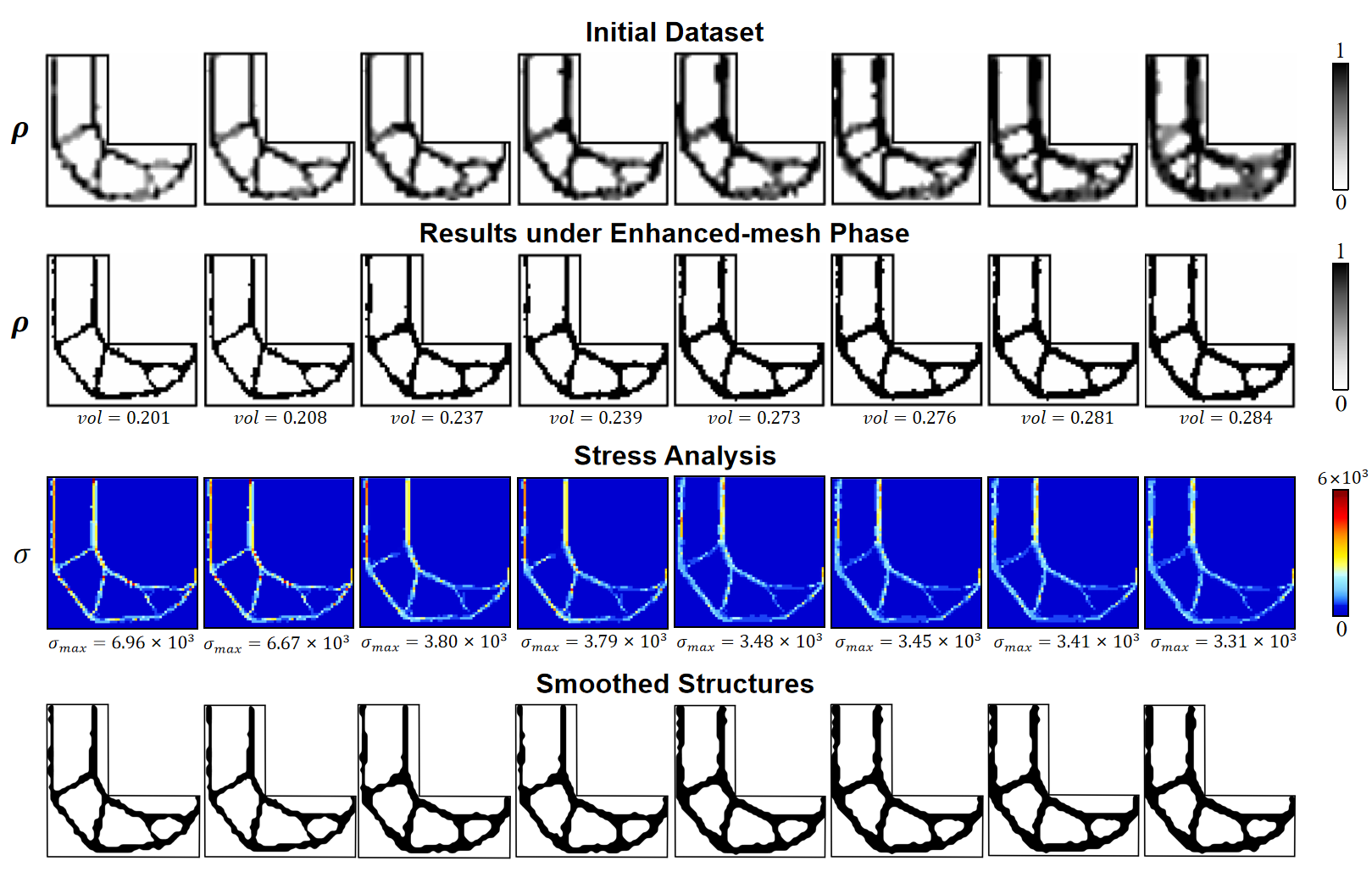}
\caption{Material distribution of elite data as well as stress analysis under enhanced-level mesh: the first row is elite data at iteration~0, the second row is elite data at iteration~125, the third row is the corresponding stress distribution of elite data at iteration~125 }
\label{Experiment 1 Mesh 2}
\end {center}
\end{figure*}

\begin{figure}[t]
\begin {center}
\includegraphics[width=0.4 \textwidth]{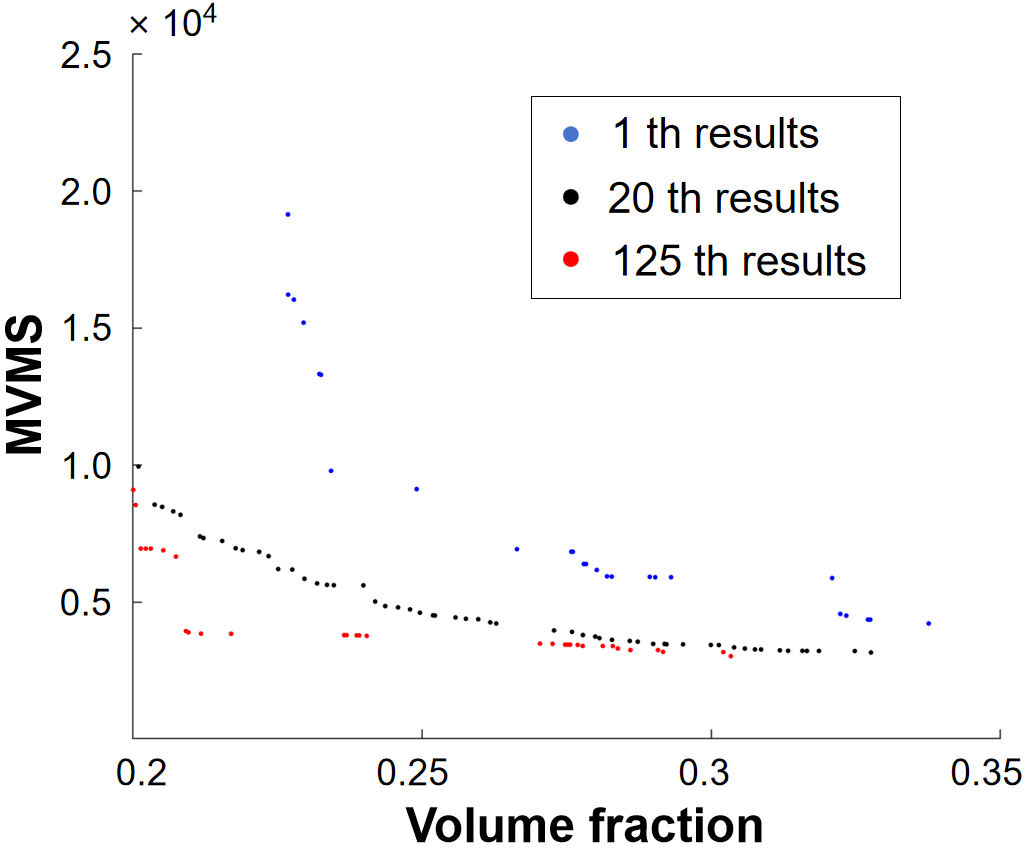}
\caption{Performances of elite material distributions: iteration~0 (blue), iteration~20 (black), and iteration~125 (red)}
\label{Improvement 1 Mesh 2}
\end {center}
\end{figure}

\begin{figure*}[t]
\begin {center}
\includegraphics[width=1 \textwidth]{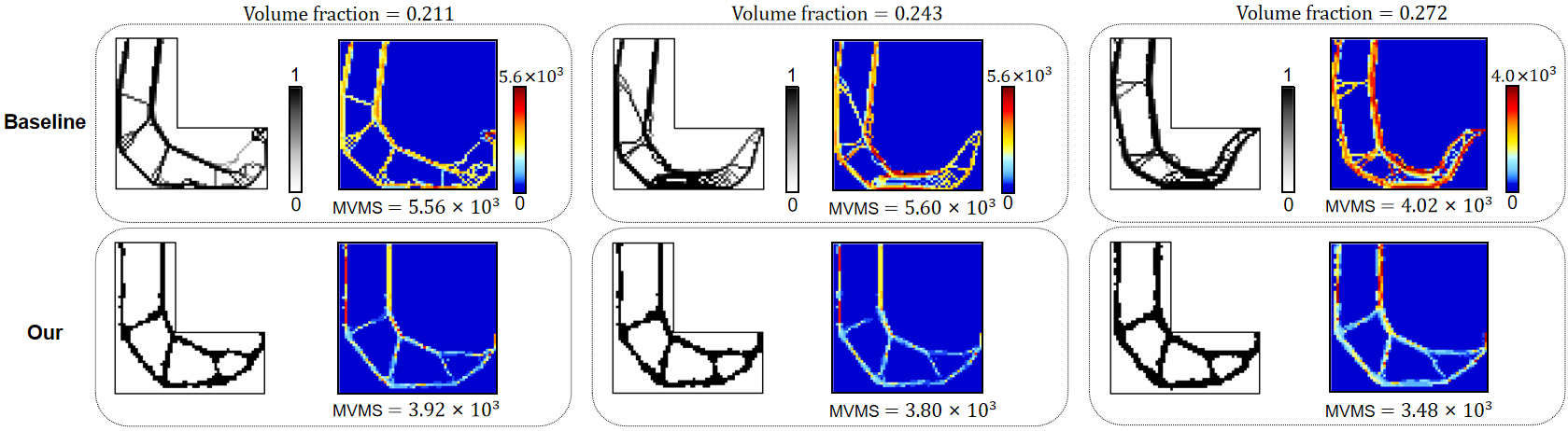}
\caption{Comparison of the final results of DDTD at enhanced-level mesh with the results of sensitivity-based TO method}
\label{Comparsion 1 Mesh 2}
\end {center}
\end{figure*}

\begin{figure*}[t]
\begin {center}
\includegraphics[width=0.9 \textwidth]{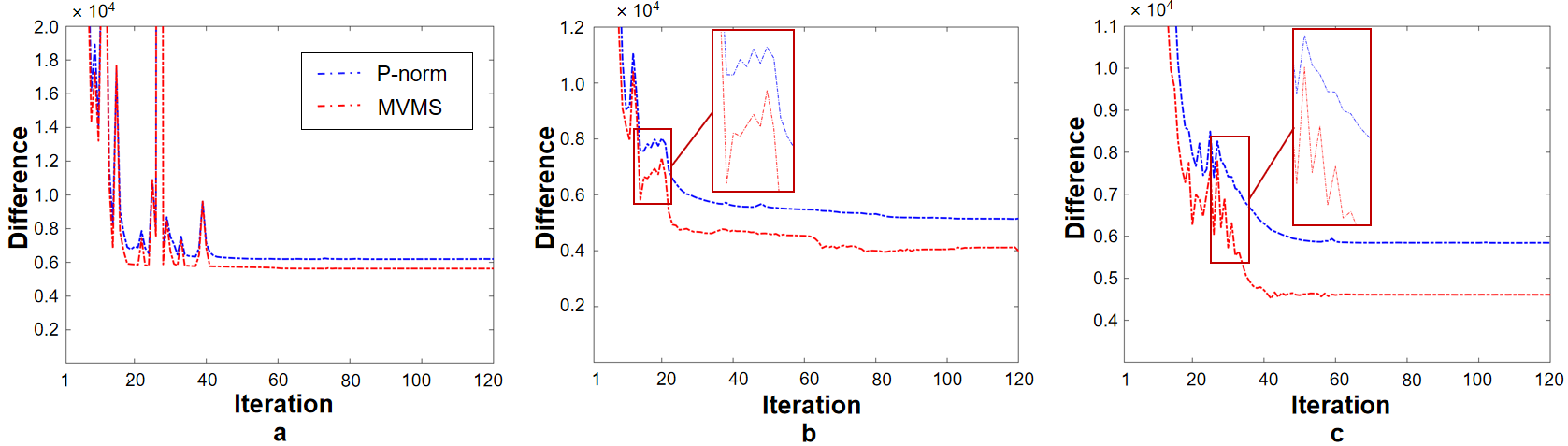}
\caption{Comparison of trends in the p-norm (approximate value) (blue line) and MVMS (true value) (red line) in results of the sensitivity-Based TO method }
\label{P norm Mesh 2}
\end {center}
\end{figure*}

\begin{figure*}[t]
\begin {center}
\includegraphics[width=1 \textwidth]{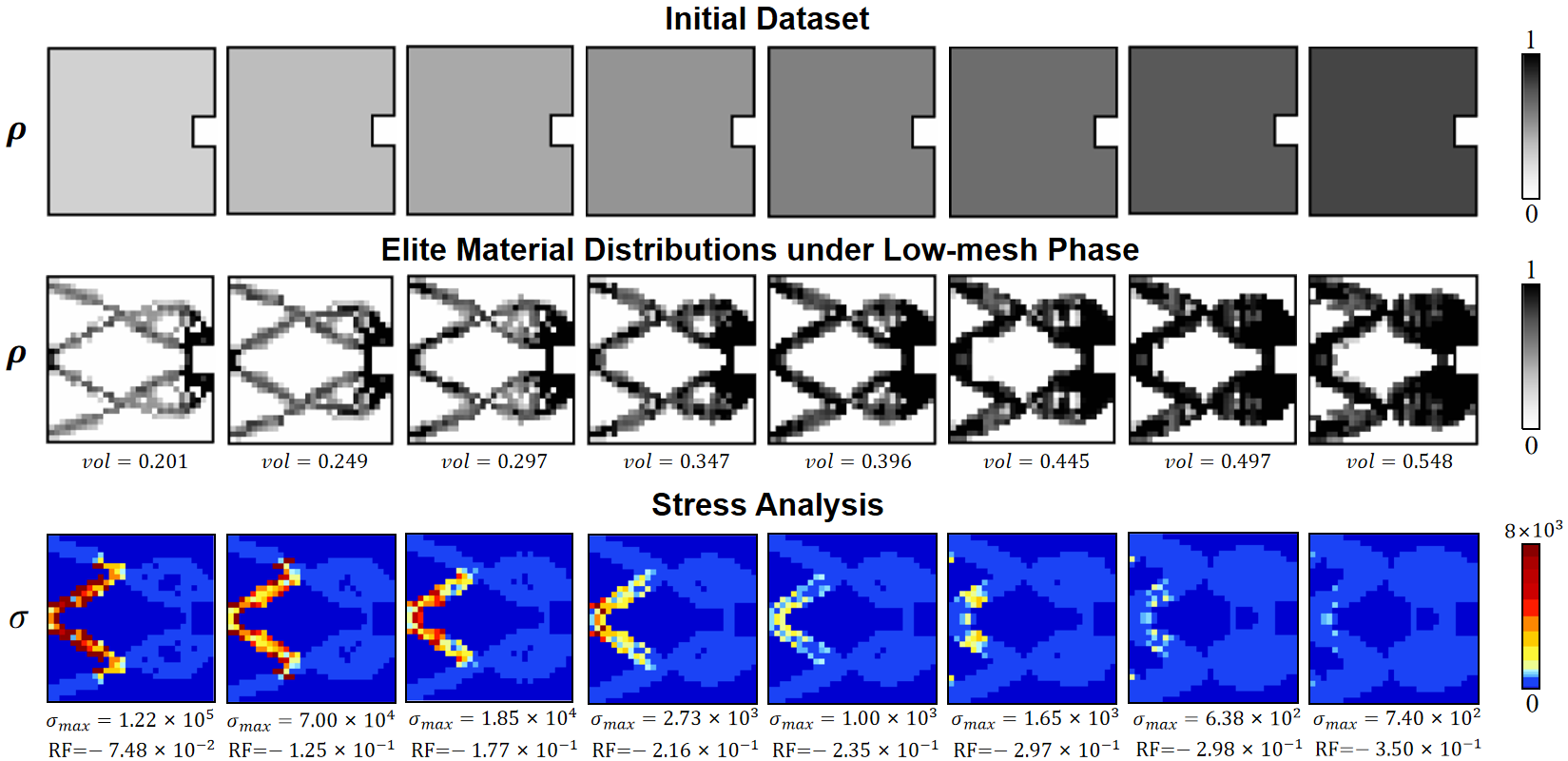}
\caption{Material distribution of elite data as well as stress analysis under low-level mesh: the first row is elite data at iteration~0, the second row is elite data at iteration~150, the third row is the corresponding stress distribution of elite data at iteration~150 }
\label{Experiment 2 Mesh 1}
\end {center}
\end{figure*}

\begin{figure*}[t]
\begin {center}
\includegraphics[width=1 \textwidth]{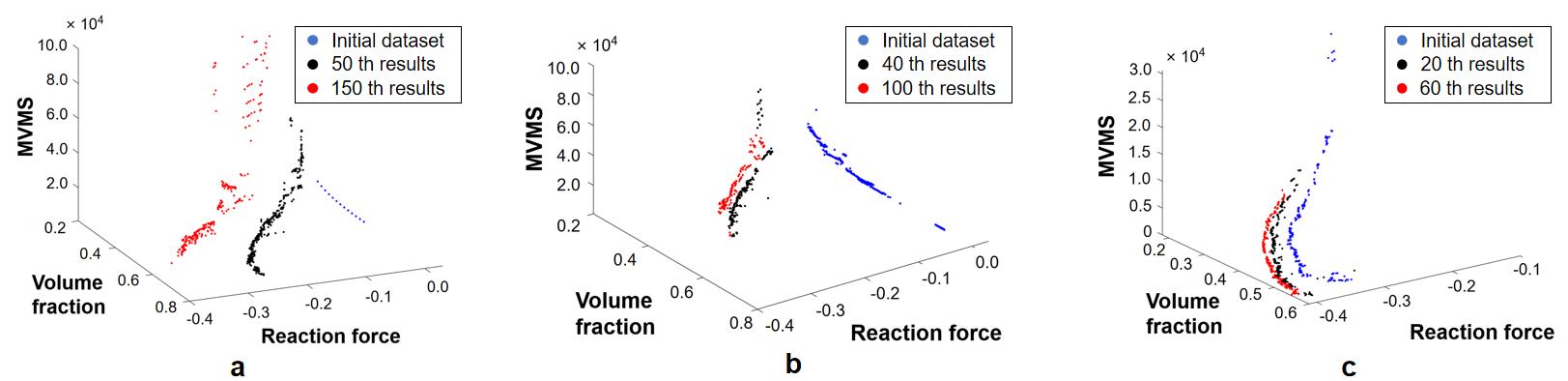}
\caption{Performances of elite material distributions in compliant mechanism design problem considering stress.}
\label{Improvement 2}
\end {center}
\end{figure*}

\begin{figure*}[t]
\begin {center}
\includegraphics[width=1 \textwidth]{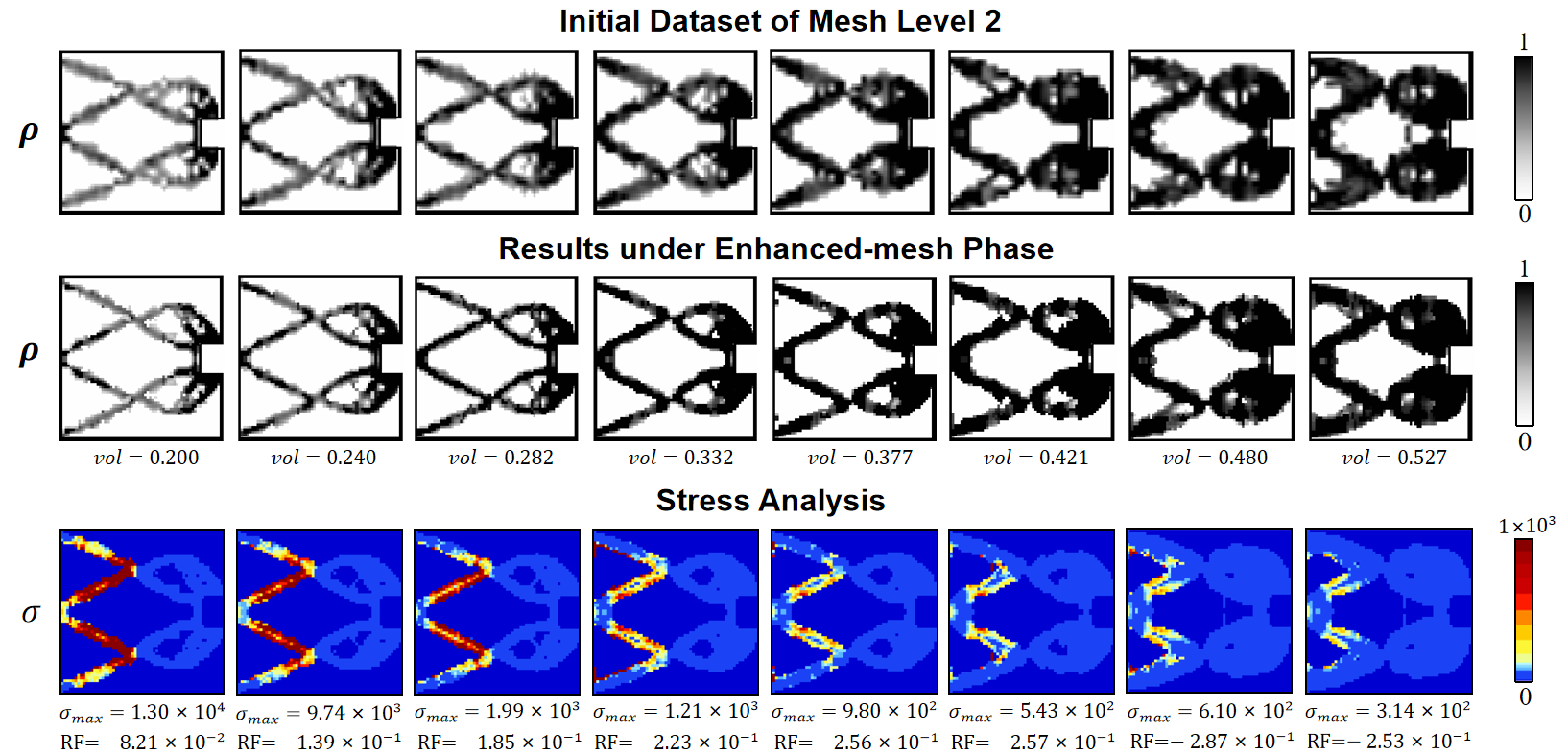}
\caption{Material distribution of elite data as well as stress analysis under enhanced-level mesh 1: the first row is elite data at iteration~0, the second row is elite data at iteration~100, the third row is the corresponding stress distribution of elite data at iteration~100}
\label{Experiment 2 Mesh 2}
\end {center}
\end{figure*}

\begin{figure*}[t]
\begin {center}
\includegraphics[width=0.85 \textwidth]{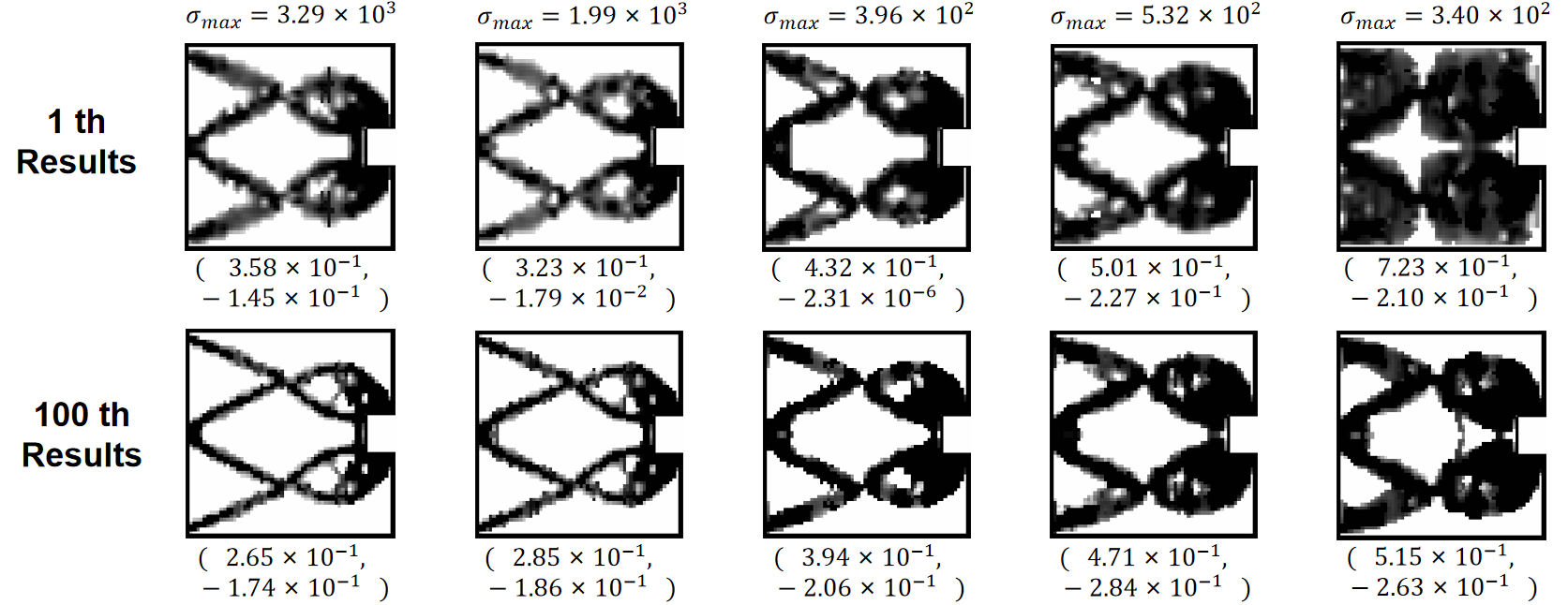}
\caption{Comparison of elite material distributions at iteration~1 and 100. Here, the left and right sides of the bracket indicate the volume fraction and reaction force values, respectively}
\label{Experiment 2 Mesh 2_comparsion}
\end {center}
\end{figure*}

\begin{figure}[t]
\begin {center}
\includegraphics[width=0.4 \textwidth]{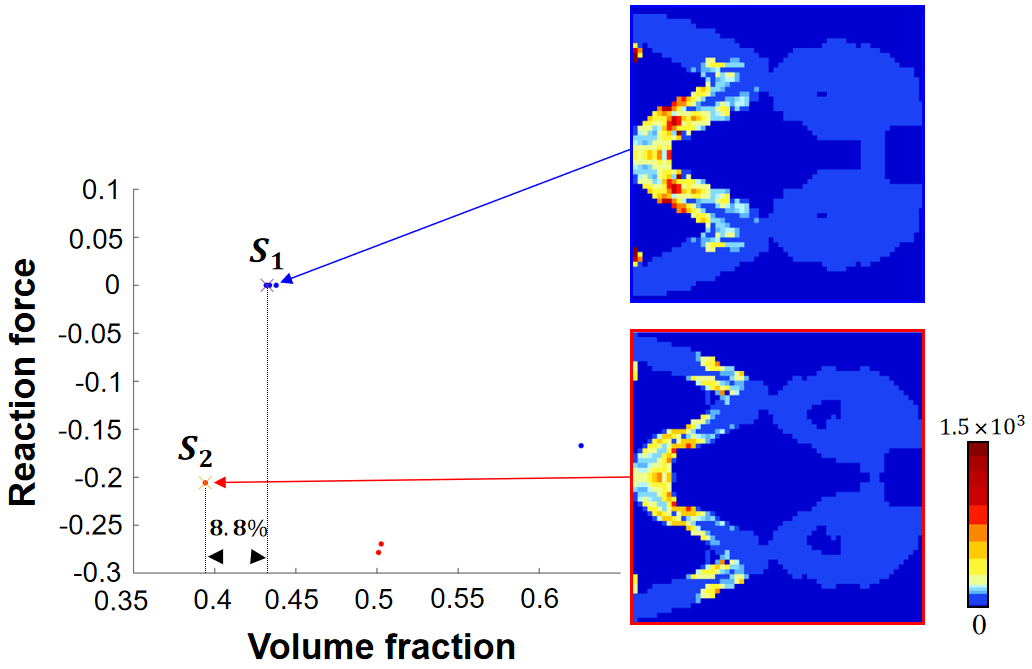}
\caption{Comparison of elite material distributions at iterations~1 (blue) and 100 (red) under MVMS $\in 3.96\times 10^{2}\pm 2$}
\label{Experiment 2 Mesh 2_2D comparsion}
\end {center}
\end{figure}

\begin{figure*}[t]
\begin {center}
\includegraphics[width=1 \textwidth]{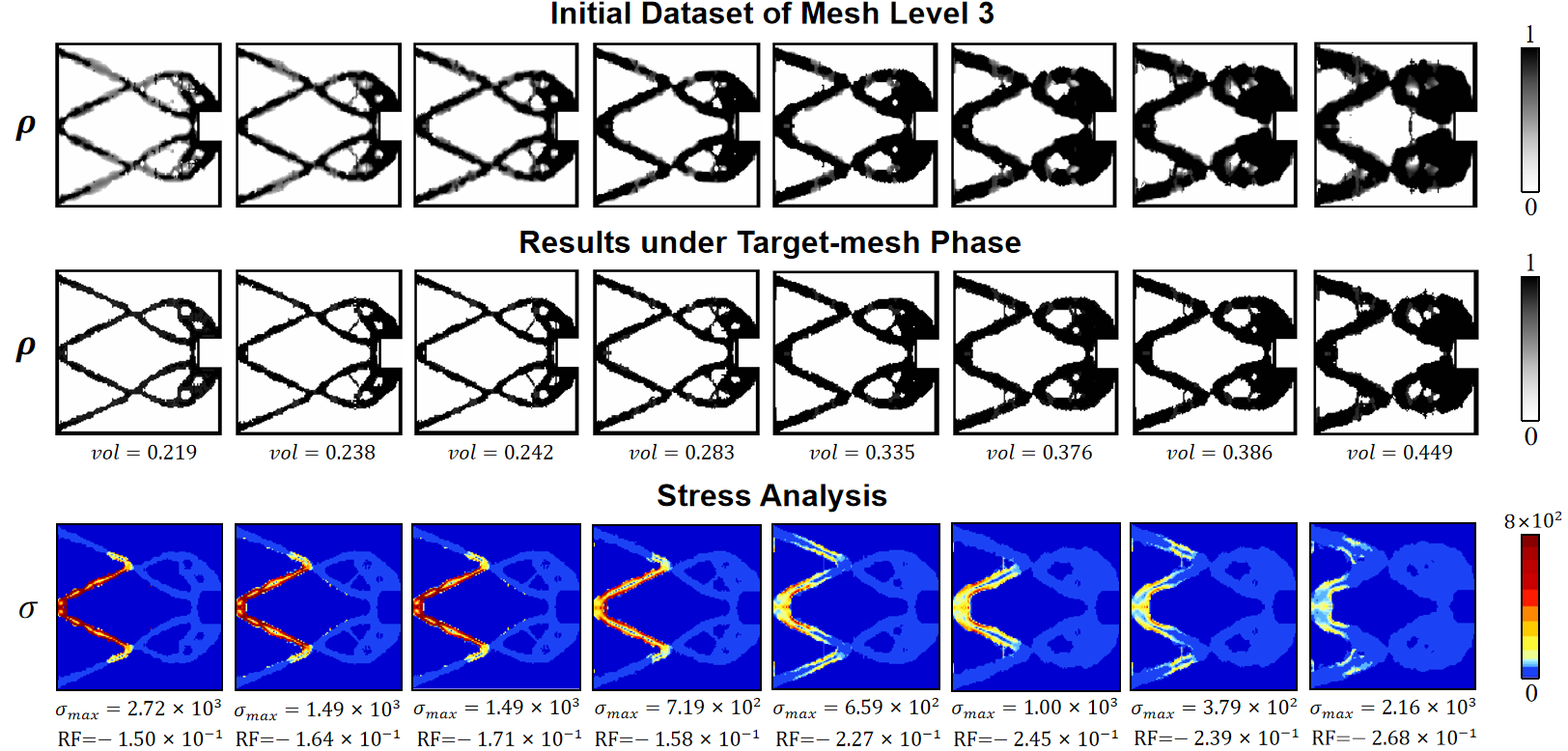}
\caption{Material distribution of elite data as well as stress analysis under enhanced-level mesh 1: the first row is elite data at iteration~0, the second row is elite data at iteration~60, the third row is the corresponding stress distribution of elite data at iteration~60}
\label{Experiment 2 Mesh 3}
\end {center}
\end{figure*}

\begin{figure*}[t]
\begin {center}
\includegraphics[width=0.85 \textwidth]{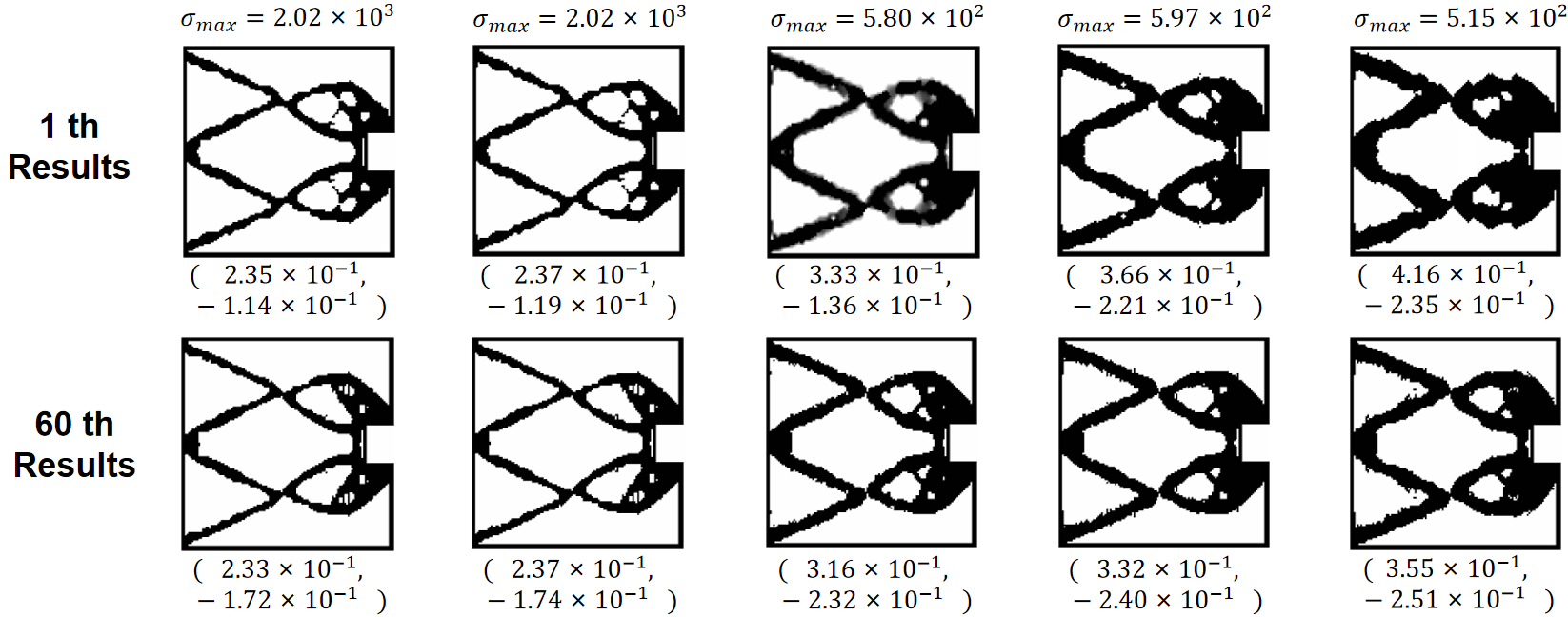}
\caption{Comparison of elite material distributions at iteration~1 and 60. Here, the left and right sides of the bracket indicate the volume fraction and reaction force values, respectively}
\label{Experiment 2 Mesh 3_comparsion}
\end {center}
\end{figure*}

\begin{figure}[t]
\begin {center}
\includegraphics[width=0.45 \textwidth]{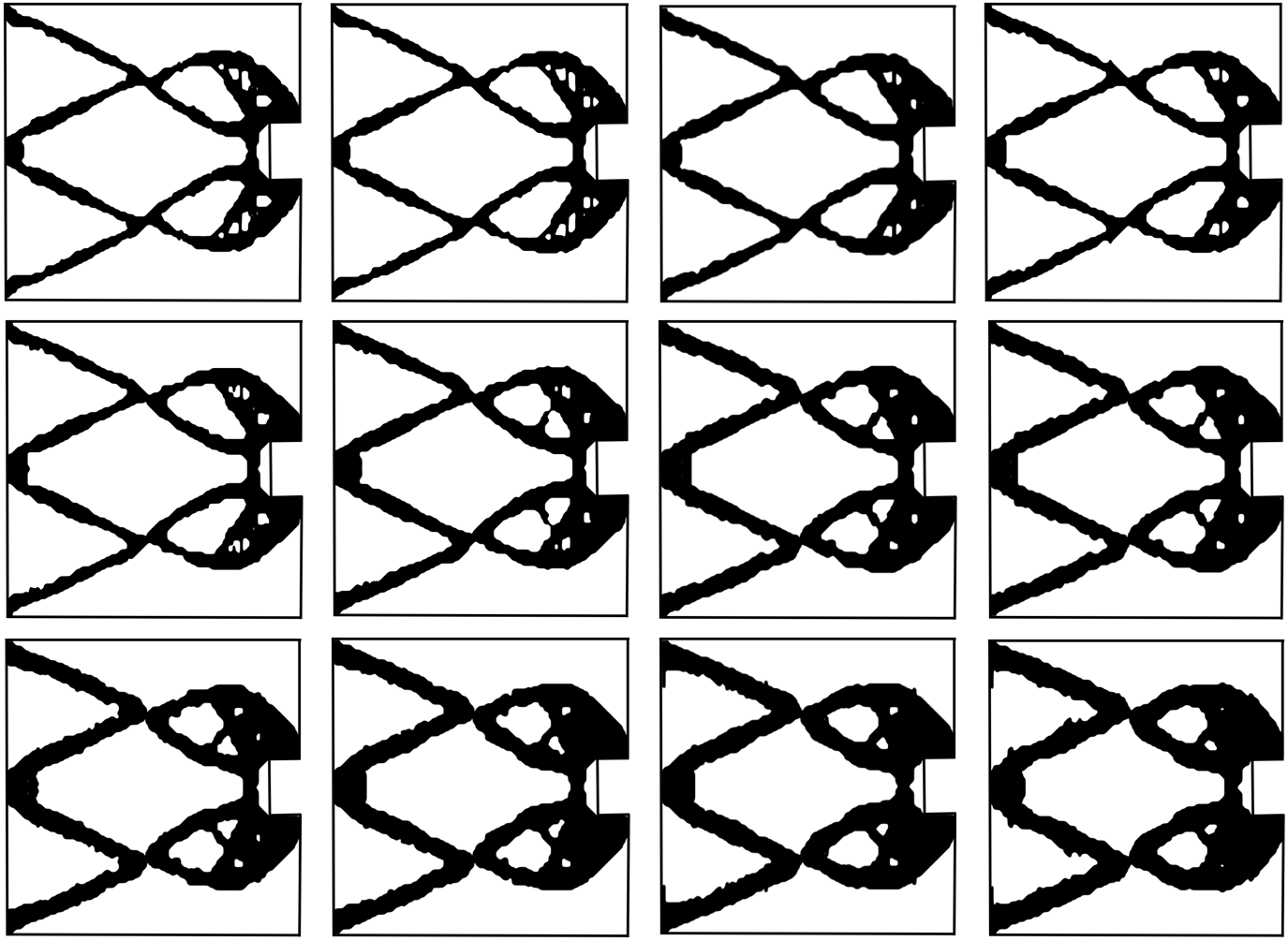}
\caption{Part of smoothed structures in the enhanced-mesh phase of $120\times 120$}
\label{Smoothed Structures2}
\end {center}
\end{figure}

\begin{figure}[t]
\begin {center}
\includegraphics[width=0.4 \textwidth]{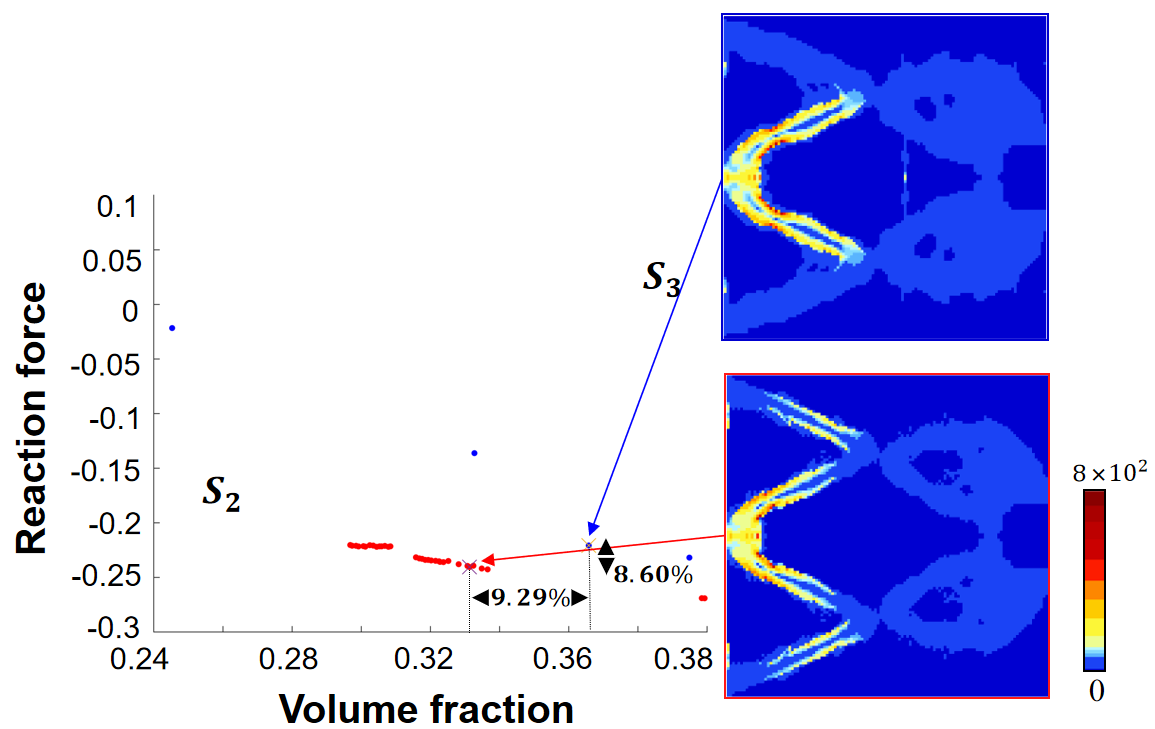}
\caption{Comparison of elite material distributions at iterations~1 (blue) and 60 (red) under MVMS $\in 5.97\times 10^{2}\pm 6$}
\label{Experiment 2 Mesh 3_2D comparsion}
\end {center}
\end{figure}

In the low-mesh phase, we firstly obtained low-performance initial material distributions by assigning a uniform constant density to all elements of one structure.
After that, we selected $11$ initial elite material distributions that satisfy the volume fraction constraint as shown in Figure~\ref{Experiment 1} first row.

Figure~\ref{Improvement 1 Mesh 1} shows how the performances of the elite material distributions are improved through the DDTD iterative process.
As shown in this figure, the elite solutions progressively migrate toward the minimization region in the objective function space, demonstrating that the performance of the elite material distributions improves through the DDTD process.
In this phase, the DDTD process, driven by the low-performance initial dataset, generates material distributions with blurred contours, resulting in a rapid decrease in the optimization objectives, i.e., the improvement of elite material distributions during the optimization process from iteration~1 to 50 is more significant compared to the improvement of elite material distributions from iteration~50 to 200.
Part of the initial dataset (elite material distributions at iteration~0), elite material distributions at iteration~200, and the corresponding stress distribution are shown in Figure~\ref{Experiment 1}.
We achieved satisfactory elite material distributions from the DDTD process driven by the low-performance dataset, demonstrating that the proposed method effectively overcomes the QID limitations faced by the original DDTD methodology, thereby significantly improving both effectiveness and generalization.

In the enhanced-mesh phase, we obtained the initial material distributions of enhanced-mesh by enhancing elite material distributions of low-mesh phase at iteration~200, as described in Sec~\ref{sec32}.
After that, we selected $400$ elite material distributions that satisfy the volume fraction constraint as initial dataset of the enhanced-mesh phase as shown in Figure~\ref{Experiment 1 Mesh 2} first row.

Figure~\ref{Improvement 1 Mesh 2} illustrates the performance enhancement of the elite material distributions achieved through the iterative DDTD process.
In this phase, the details of the material distribution are gradually refined based on the approximate topological shape obtained during the low-mesh phase. 
This process continues until convergence to the optimal material distributions is achieved, resulting in slighter changes in the optimization objective function.
Part of the initial dataset (elite material distributions at iteration~0), elite material distributions at iteration~125, and the corresponding stress distribution are shown in Figure~\ref{Experiment 1 Mesh 2}.
As described in Sec~\ref{sec32}, we update the generated data using the heavside function in this phase to reduce the number of greyscale elements, it results in a clear boundary for the material distributions.
To improve the manufacturability of the jagged results in the second row of Figure~\ref{Experiment 1 Mesh 2}, we use MATLAB's contour function to obtain the $\rho=0.5$ level boundaries and fill $\rho>0.5$ area  as the final result as shown in the fourth row of Figure~\ref{Experiment 1 Mesh 2}.
The average time per iteration for the overall process (including low-mesh and enhanced mesh phase) is approximately $26.8$ minutes.

Here, we chose three structures from elite material distributions of enhanced-mesh phase at iteration~125 to compare with the sensitivity-based TO method (\cite{deng2021efficient}) to demonstrate the effectiveness of the proposed approach and to explain the potential limitations that may exist in the sensitivity-based TO method under stress-related optimization problems.
As shown in Figure~\ref{Comparsion 1 Mesh 2}, the first row shows the results of the sensitivity-based TO optimization method, serving as baseline, while the second row displays the results of the DDTD methodology. 
To ensure a fair comparison, both methods were solved under as similar conditions as possible, including design domain and boundary conditions, maximum number of iterations, filter parameters, etc.
As shown in this figure, the three sets of comparison data were evaluated under similar volume fractions (difference within $1\times 10^{-3}$). 
It can be observed that the DDTD results yield lower MVMS values (with DDTD results being 3.92, 3.80 ,and 3.48 $\times 10^{3}$, compared to the baseline values of 5.56, 5.60 ,and 4.02 $\times 10^{3}$), demonstrating that DDTD can achieve better material distributions than the sensitivity-based TO method.

As previously mentioned, sensitivity-based TO methods may exist the potential limitations when minimizing the maximum von Mises stress due to the difficulties in directly computing its sensitivity, which can result in discontinuities and numerical instability. Consequently, the use of the p-norm way to transform local maximum stress into a globally smoothed stress measure improves the solvability of sensitivity calculations; however, the discrepancy between this approximation and the true values may introduce some computational bias.
Typically, sensitivity-based TO methods employ the following p-norm function to approximate the maximum von Mises stress (MVMS):
\begin{equation}
\label{pnorm}
\max \left(\sigma_{i}^{\text {mise }}\right) \approx \sigma_{p \text {-norm }}=\left(\sum_{j=1}^{n_{e}}\left(\sigma_{i j}^{\text {mise }}\right)^p\right)^{1 / p}
\end{equation}
where, $\max \left(\sigma_{i}^{\text {mise }}\right)$ is the MVMS value in the $i$-th structure, i.e., the true value.
$\sigma_{p \text {-norm }}$ is the equivalent MVMS value obtained by p-norm approximation and is used as a substitute for the MVMS , i.e., approximate value.
$\sigma_{i j}^{\text {mise }}$ is the VMS value of the $j$-th element of the $i$-th structure.
$q$ is a constant value, usually $q=10$.

Figure~\ref{P norm Mesh 2} shows the trend of the p-norm (approximate value) and MVMS (true value) corresponding to the results of the sensitivity-based TO process in the first row of Figure~\ref{Comparsion 1 Mesh 2} during the optimization process.
We ignore the sharp fluctuations at the beginning of the iteration due to the difficulty of observation.
We observe that the values and trends of the p-norm and MVMS during the iteration process cannot be strictly equated to one another and may even exhibit opposing changing trends (such as the neighborhood of $30$-th iterations in (c)).
Instead, our method evaluates the structure directly using MVMS, which can be free from the potential limitations of using this approximation way (P-norm).

\subsection{Numerical example 2} \label{exam2}

To further validate the effectiveness of the proposed methodology, we chose compliant mechanism design problem considering stress as shown in Figure~\ref{des}(b).
We establish three levels of mesh, meaning that the mesh needs to be enhanced twice.
The discretization of the low-mesh phase and two enhanced mesh phase are $30\times30$, $60\times60$, and $120\times120$, respectively.
In this optimization problem, the number of design variables is reduced by half compared to the original design domain, owing to the symmetric boundary conditions along the $x$ plane, i.e., $450$, $1800$, and $7200$, respectively.
The maximum number of iterations for the low-mesh phase and two enhanced mesh phase are $150$, $100$, and $60$ respectively.
Furthermore, we imposed a volume fraction constraint ($J_2\in [0.2, 0.8]$) to ensure that the results fall within the target range.
For additional problem settings, a horizontal load of $150$ is applied at the input port, an artificial spring with a stiffness of $10$ is attached to the output port, and the displacement is constrained at the fixed end.
Young's modulus and Poisson's ratio are set to $1$ and $0.3$, respectively.

In the low-mesh phase, we firstly obtained low-performance initial material distributions by assigning a uniform constant density to all elements of one structure.
Subsequently, we selected $13$ initial elite material distributions that satisfy the volume fraction constraint as shown in Figure~\ref{Experiment 2 Mesh 1} first row.

Figure~\ref{Improvement 2}(a) demonstrates the improvement in the performance of elite material distributions through the iterative DDTD process.
As shown in the figure, the elite solutions progressively move toward the minimization region in the three-dimensional objective function space, indicating that the DDTD process effectively enhances the performance of the elite material distributions.
In this phase, the DDTD process, driven by the low-performance initial dataset, generates material distributions with blurred contours, resulting in a rapid decrease in the optimization objective function.
From the iteration~1 to the 50, and then to the 150, the improvement of elite material distributions is consistently significant throughout the entire process.
Part of the initial dataset (elite material distributions at iteration~0), elite material distributions at iteration~150, and the corresponding stress distribution are shown in Figure~\ref{Experiment 2 Mesh 1}.
We achieved satisfactory elite material distributions from the DDTD process driven by the low-performance dataset, demonstrating that the proposed method effectively overcomes the QID limitations faced by the original DDTD methodology, thereby significantly improving both performance and generalizability.

In the enhanced-mesh phase of $60\times 60$, we obtained the initial material distributions of enhanced-mesh by enhancing elite material distributions of low-mesh phase at iteration~150.
After that, we selected $400$ elite material distributions that satisfy the volume fraction constraint as initial dataset of the enhanced-mesh phase.

Figure~\ref{Improvement 2}(b) illustrates the improvement in the performance of elite material distributions through the iterative DDTD process.
As shown in the figure, the elite solutions gradually converge towards the minimization region in the objective function space, demonstrating that the DDTD process effectively enhances the performance of the elite material distributions.
In this phase, the details of the material distribution are gradually refined based on the approximate topological shape obtained during the low-mesh phase. This process continues until convergence to the better material distributions is achieved, resulting in significant changes in the optimization objective function.
Compared to the significant improvement in elite material distribution from iteration~1 to 40, the rate of improvement from iteration~40 to 100 decreases substantially.
Part of the initial dataset (elite material distributions at iteration~0), elite material distributions at iteration~100, and the corresponding stress distribution are shown in Figure~\ref{Experiment 2 Mesh 2}.

We here chose five structures from elite material distributions of low-mesh phase at iteration~1 and iteration~100 as shown in Figure~\ref{Experiment 2 Mesh 2_comparsion} to compare the changes in material distribution shapes, topologies, and optimization objective values during the iteration process to further demonstrate the effectiveness of the proposed approach.
Since this compliant mechanism design problem contains three optimization objectives (volume fraction, reaction force, MVMS), we chose three pairs of results under almost same MVMS (difference within $2$) for straightforward comparison.
As we have seen, the structural performance of the elite material distributions at iteration~100 is significantly better than the structural performance of the elite material distributions at iteration~1 (i.e., having smaller objective function values in both volume fraction and reaction force).
Figure~\ref{Experiment 2 Mesh 2_2D comparsion} shows the position of the third comparison data from Figure~\ref{Experiment 2 Mesh 2_comparsion} within the improvement space. 
The blue dot represents the elite material distribution at iteration~1, while the red dot represents the elite material distribution for similar MVMS $\in 3.96\times 10^{2}\pm 2$ at iteration iteration~100. 
We can observe a significant improvement in the performance of the elite material distribution, with volume fraction reducing from $4.32\times 10^{-1}$ to $3.94\times 10^{-1}$ (a 8.8\% decrease), and reaction force decreasing from $-2.31\times 10^{-6}$ to $-2.06\times 10^{-1}$.

In the enhanced-mesh phase of $120\times 120$, we obtained the initial material distributions of enhanced-mesh by enhancing elite material distributions of the enhanced-mesh phase of $60\times 60$ at iteration~100.
After that, we selected $400$ elite material distributions that satisfy the volume fraction constraint as initial dataset of the enhanced-mesh phase.
Figure~\ref{Improvement 2}(c) illustrates the improvement in the performance of elite material distributions through the iterative DDTD process.
In this phase, the details of the material distribution, characterized by a certain topological shape obtained from the last enhanced-mesh phase, are further refined. 
Due to the sufficiently good performance of the initial data, there are no significant changes in the numerical values of the optimization objective.
Part of the initial dataset (elite material distributions at iteration~0), elite material distributions at iteration~60, and the corresponding stress distribution are shown in Figure~\ref{Experiment 2 Mesh 3}.
We update the generated data using the heavside function in this phase to reduce the number of grayscale elements, it results in a clear boundary for the material distributions.
To improve the manufacturability of the jagged results in the second row of Figure~\ref{Experiment 2 Mesh 3}, 
the $\rho=0.5$ level boundaries are calculated and used as the final result, and a part of the smoothed final results are shown in Figure~\ref{Smoothed Structures2}.
The average time per iteration for the overall process (including low-mesh and enhanced mesh phase) is approximately $25.2$ minutes.

We here still chose five structures from elite material distributions of low-mesh phase at iteration~1 and iteration~60 as shown in Figure~\ref{Experiment 2 Mesh 3_comparsion} to compare the changes in material distribution shapes, topologies, and optimization objective values.
We chose five pairs of results under almost same MVMS (difference within $6$) for straightforward comparison.
As observed, the structural performance of the elite material distributions at iteration~60 is better than the structural performance of the elite material distributions at iteration~1.
Figure~\ref{Experiment 2 Mesh 3_2D comparsion} shows the position of the fifth comparison data from Figure~\ref{Experiment 2 Mesh 3_comparsion} within the improvement space. 
The blue dot represents the elite material distribution at iteration~1, while the red dot represents the elite material distribution for similar MVMS $\in 5.97\times 10^{2}\pm 6$ at iteration iteration~60. 
We can observe a significant improvement in the performance of the elite material distribution, with volume fraction reducing from $3.66\times 10^{-1}$ to $3.32\times 10^{-1}$ (a 9.29\% decrease), and reaction force decreasing from $-2.21\times 10^{-1}$ to $-2.40\times 10^{-1}$ (a 8.60\% decrease).

\section{Conclusions}\label{sec5}

In this paper, we propose a multi-level mesh data-driven topology design methodology with correlation-based mutation module, which effectively addresses the limitation in original DDTD methodology, where results are highly influenced by the quality of the initial dataset. 
This significantly improves the effectiveness and generalizability of the original DDTD methodology and mitigates its over-reliance on machine learning technology. 
The multi-level mesh introduced in this paper reduces the computational cost of original DDTD methodology, improves computational efficiency, and lowers the overhead caused by maintaining high DOF structural representation.
By comparing the proposed method with conventional sensitivity-based TO method, we have validated its advantages in solving stress-related strongly nonlinear problems, as well as highlighted the limitations and challenges faced by sensitivity-based TO methods when dealing with stress-related strongly nonlinear optimization problems.
In future work, we aim to further expand the application of proposed method to more complex design problems, e.g. 3D strongly nonlinear problems, providing a more robust tool for structural design.
Besides, we will explore the feasibility of using machine learning techniques to directly predict structural performance for the purpose of reducing the computational cost of DDTD.
\\

\section*{Appendix} 

The convergence criterion of the proposed method is based on the area outside of the elite solutions within the objective space.
If the change in the area is below a predetermined threshold $\epsilon$ (usually $1\times10^{-6}$) for 15 consecutive iterations (convergence criterion), or if the iteration count reaches the maximum number of iteration, the data generation process is terminated.

\begin{figure}[t]
\begin {center}
\includegraphics[width=0.5 \textwidth]{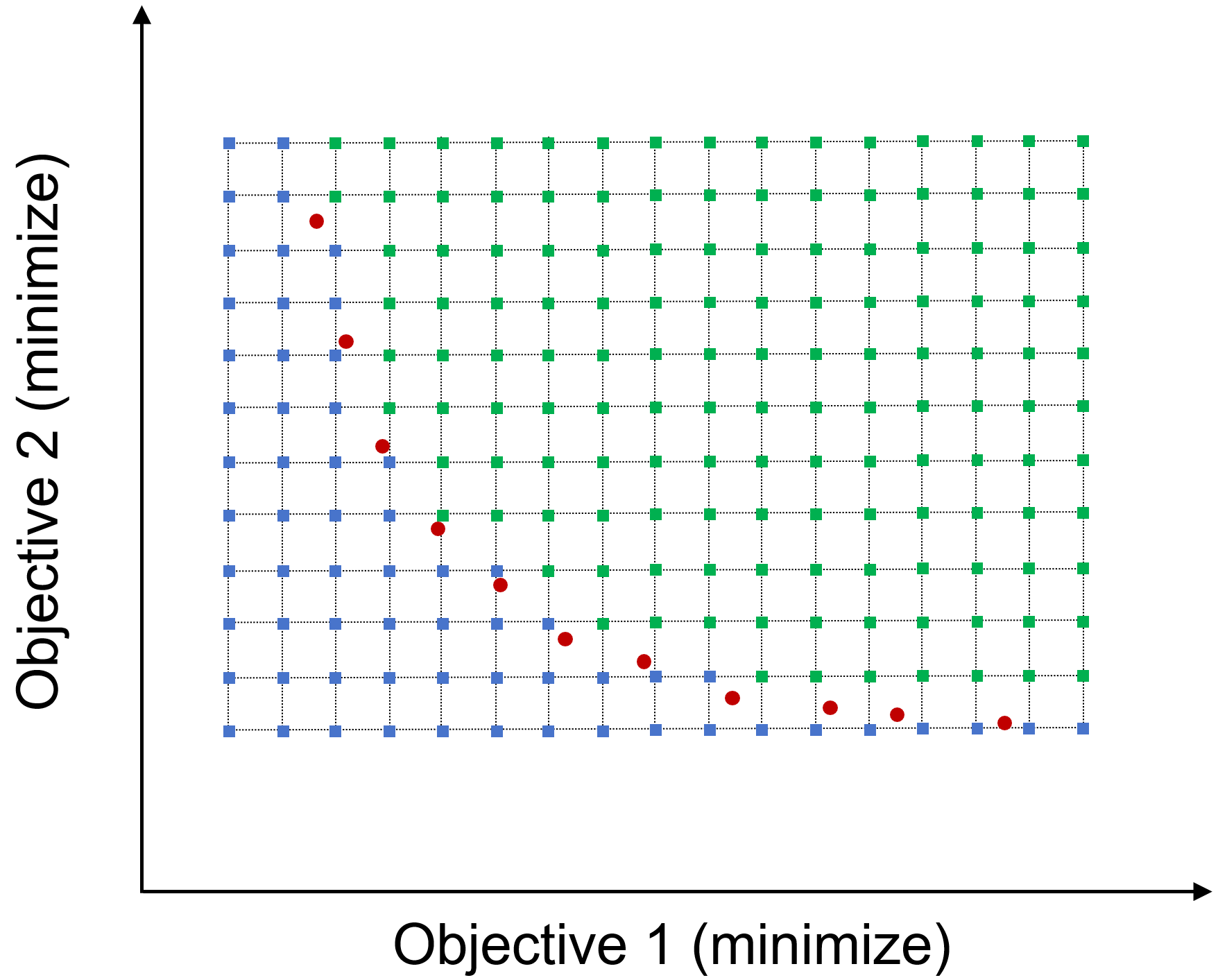}
\caption{An example of the fixed grid points for approximately computing
the area outside elite solutions colored with red}
\label{area}
\end {center}
\end{figure}

The calculation of the area outside the elite solutions is approximated by counting the fixed grid points that are not dominated by the elite solutions \cite{yamasaki2021data}.
For normalization purposes, the ratio of the number of the fixed grid points that are not dominated to the number of all the fixed grid points is taken as the area outside the elite solutions in this paper.
Specifically, for multi-objectives optimization problem, a grid point $p_i$ is not dominated by elite solutions $\mathcal{S} = \{s_1, s_2, ..., s_e\}$ ($e$ is the number of the elite solutions) with objective values $\left[ J_1, J_2, ..., J_{N_{obj}} \right]$ if it satisfies the following conditions:
\begin{equation*}
\begin{array}{l}
    \forall k \in \{1, ..., {N_{obj}}\}, \quad J_k(p_i) \leq J_k(s_j),\\ \\
    \exists k \text{ such that } J_k(p_i) < J_k(s_j),
\end{array}
\end{equation*}
where, $J_k(p_i)$ is the $k$-th objective value of the fixed grid point $p_i$.
For example, as shown in Figure~\ref{area}, the green grid points
are dominated by the blue elite solutions, and the blue
grid points are not dominated. 
The area outside the red elite solutions is approximately computed by counting the ratio of the number of blue grid points to the number of all the fixed grid points (blue and green).

Thus, given a set of elite solutions (e.g., elite data in each iteration), the area outside the elite solutions can be computed based on the fixed grid points.
In this paper, the fixed grid points are the same for each phase, and its coordinates are related to the maximum and minimum value of the optimization objectives of the elite data at initial iteration.

\section*{Statements and Declarations}

\textbf{Author contributions}: The first author proposed the idea, implemented the code, conducted the experiments, and drafted the manuscript. The second author supervised the implementation details, contributed to the experimental design, provided guidance on manuscript writing, and verified the data. Both authors reviewed and approved the final version of the manuscript. \\

\noindent
\textbf{Data Availability}: The data supporting the findings of this study are available upon reasonable request from the corresponding author. \\

\noindent\textbf{Ethics approval and Consent to participate}: Not applicable. \\

\noindent\textbf{Conflict of interest}: The authors declare that they have no conflict of interest. \\

\noindent
\textbf{Funding}: Non. \\

\noindent\textbf{Replication of results}: The necessary information for a replication of the results are presented in the manuscript.
Interested readers may contact the corresponding author for further details regarding the implementation.

\bibliography{sn-bibliography}



\bibliographystyle{plain}  

\end{document}